\documentclass{article}

\PassOptionsToPackage{numbers, compress}{natbib}
\usepackage[preprint]{neurips_2026}

\usepackage[utf8]{inputenc} % allow utf-8 input
\usepackage[T1]{fontenc}    % use 8-bit T1 fonts
\usepackage{hyperref}       % hyperlinks
\usepackage{url}            % simple URL typesetting
\usepackage{booktabs}       % professional-quality tables
\usepackage{amsfonts}       % blackboard math symbols
\usepackage{nicefrac}       % compact symbols for 1/2, etc.
\usepackage{microtype}      % microtypography
\usepackage{xcolor}         % colors
\usepackage{amsmath}
\usepackage{graphicx}
\usepackage[most]{tcolorbox}                                          
\usepackage{enumitem}
\usepackage{caption}                       
\usepackage{etoc}
\tcbuselibrary{skins,breakable}   
\newcommand{\firstpagefootnote}[1]{%
  \begingroup
  \renewcommand\thefootnote{}%
  \footnotetext{\hspace{-1.8em}#1}%
  \endgroup
}
\title{VCap: Hypergeometric Rewards \\ for Weak-to-Strong Visual Captioning}

\author{%
  \textbf{Xingyu Lu}$^{1,\dagger}$ \quad
  \textbf{Jinpeng Wang}$^{2,\ddagger}$ \quad
  \textbf{Yi-Fan Zhang}$^{3}$ \quad
  \textbf{Yankai Yang}$^{4}$ \quad
  \textbf{Yancheng Long}$^{4}$\\
  \textbf{Yiyang Fan}$^{4}$ \quad
  \textbf{Xuanyu Zheng}$^{4}$ \quad
  \textbf{Haonan Fan}$^{4}$ \quad
  \textbf{Kaiyu Jiang}$^{4}$ \quad
  \textbf{Tianke Zhang}$^{4}$\\
  \textbf{Changyi Liu}$^{4}$ \quad
  \textbf{Bin Wen}$^{4}$ \quad
  \textbf{Fan Yang}$^{4}$ \quad
  \textbf{Tingting Gao}$^{4}$ \quad
  \textbf{Han Li}$^{4}$ \quad
  \textbf{Chun Yuan}$^{1,\ddagger}$\\
  \normalfont
  $^{1}$Tsinghua Shenzhen International Graduate School \quad
  $^{2}$Harbin Institute of Technology, Shenzhen\\
  $^{3}$Chinese Academy of Sciences \quad
  $^{4}$Kuaishou Technology\\
  \footnotesize $^\dagger$ Project leader. \quad
  $^\ddagger$ Corresponding authors: Jinpeng Wang and Chun Yuan.%
}

\begin{document}

\maketitle
\firstpagefootnote{Email: \texttt{xylu18@gmail.com}.}

\begin{abstract}
Visual captioning requires models to capture visual content faithfully while minimizing both omission and hallucination.
As the dominant paradigm for captioning, MLLMs have achieved strong performance through scaling and high-quality data. Recently, RL has emerged as a key route to driving MLLMs toward higher precision and broader coverage, however, existing reward designs for captioning fail to provide fine-grained and reliable signals for factual verification, limiting their effectiveness.
To address this, we propose \textbf{VCap}, a Witness-Adjudicator reward that pairs the reference caption (a \emph{witness}) with the visual signal (an \emph{adjudicator}).
By explicitly verifying factual consistency between the reference and policy-generated captions grounded in the visual signal, VCap delivers a reward signal with hypergeometric-distribution-level precision for caption quality verification. This design enables effective learning even from imperfect references, facilitating weak-to-strong generalization in RL training.
In our experiments, an 8B model trained with VCap outperforms open- and closed-source SOTA models on multiple image and video captioning benchmarks. Human evaluation further confirms its strong alignment with factual correctness. Additionally, VCap improves MLLM perceptual capability, generalizes across tasks, and surpasses best-of-N distillation, challenging prior assumptions about RLVR.
\end{abstract}

\section{Introduction}
\label{sec:intro}

\begin{figure}[h]
\centering
\includegraphics[width=\linewidth]{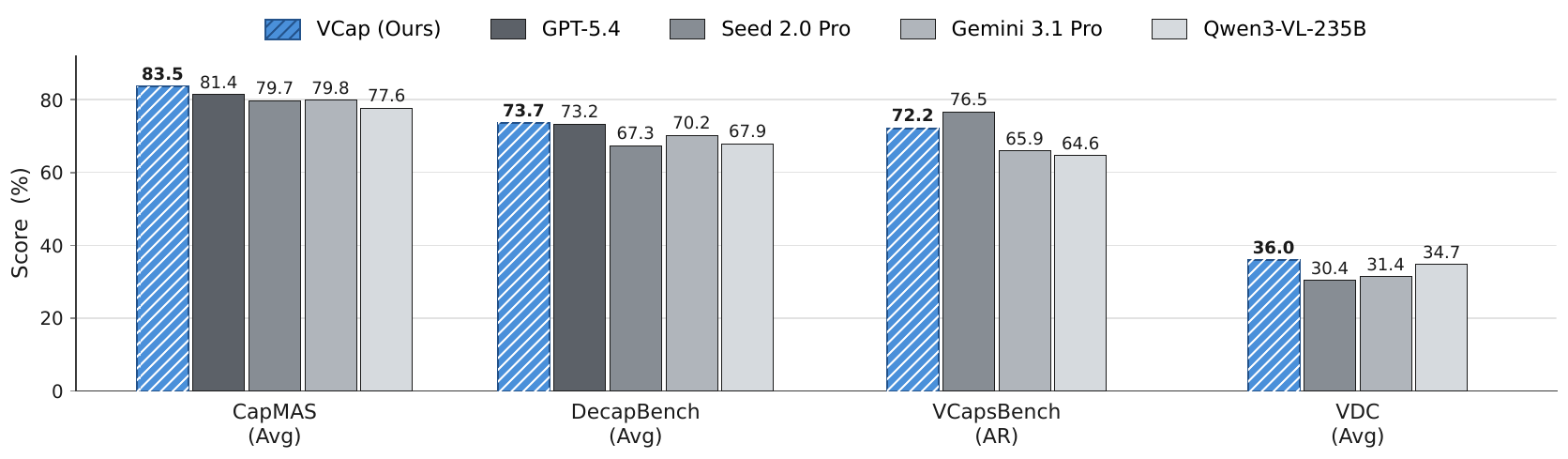}
\caption{VCap (8B) vs. frontier models across visual captioning benchmarks.}
\label{fig:overview}
\end{figure}
Visual captioning translates visual perception into natural language. A reliable captioner should express salient visual facts, including objects, attributes, relationships, and events, while avoiding both omissions and fabrications. At the fact level, caption quality is governed by two complementary dimensions: \textbf{Correctness}, whether each stated fact is grounded in the visual signal, and \textbf{Completeness}, whether the important visual content is sufficiently covered. The impact extends beyond captioning itself: faithful captions serve as supervision, intermediate representations, and synthetic data for cross-modal alignment, retrieval, agent reasoning, and downstream multimodal training, where factual errors can silently propagate.

Existing efforts to improve caption quality follow three main routes: scaling models and pretraining data, curating and distilling stronger reference captions, and optimizing captioners with reinforcement learning (RL). The first two approaches share a common ceiling: even carefully curated data inevitably propagate omissions and hallucinations into the model. RL, in principle, offers a way beyond this limitation. Dominant reward designs, VQA-based~\citep{caprl}, VLM-as-judge~\citep{xiong2024llavacritic}, and imitation-style rewards~\citep{wang2024mdpo,xie2024vdpo} (e.g., CIDEr, DPO), all constrain reference signals to a single role: a limited question pool, a judge’s ability, or an explicit imitation target. These designs can provide useful signals, but they struggle to jointly supervise \emph{both} correctness and completeness at fine granularity. The core obstacle is that the complete set of visual facts is unavailable in practice: no reference enumerates it, and no reward proxy fully recovers it.

To address this limitation, we introduce \textbf{VCap}, which separates the roles of textual reference and visual evidence in reward modeling. VCap treats the reference caption not as a target to imitate, but as partial evidence about the visual input. The reference caption acts as a stochastic \textbf{witness}: to expose a subset of visual facts. This witness plays a dual role. As a \textbf{recall anchor}, it requires the policy caption to cover the facts it mentions. As an \textbf{error detector}, it helps identify hallucinated policy claims when they collide with witnessed facts under visual verification. The visual input plays the complementary role of an \textbf{adjudicator}: it verifies matches and mismatches at the witness-activated slots, but does not serve as an oracle that enumerates all facts in the scene.

This role separation admits a closed-form analysis. Let $\mathcal{F}$ denote the latent visual fact set with $|\mathcal{F}|=N$, $R$ the reference fact subset with $|R|=m$, and $\Phi(y)=C\cup E$ the policy fact set, where $|C|=c$ are correct facts and $|E|=n-c$ are hallucinated facts. Treating $R$ as a stochastic subset of $\mathcal{F}$, VCap yields hypergeometric reward factors for completeness and correctness:
\begin{equation}
  P_{\text{comp}} \;=\; \prod_{i=0}^{m-1}\frac{c - i}{N - i},
  \qquad
  P_{\text{corr}} \;=\; \prod_{j=0}^{n-c-1}\frac{N - m - j}{N - j}.
  \label{eq:ppass}
\end{equation}
The first term increases as the policy covers more correct facts, while the second decreases as hallucinations grow. Both are jointly optimized when the policy approaches the image-information ceiling, i.e., $c \to N$ and $n-c \to 0$. Crucially, the location of this optimum does not depend on the reference size $m$: smaller or imperfect references change the steepness of the supervision landscape, but not its optimum. We call this property \emph{two-axis weak-to-strong generalization}. Unlike imitation-style rewards, whose optimum is tied to the reference caption itself, VCap uses the reference as sampled probes of the latent fact space, allowing weak references to guide stronger captioners.

We evaluate these predictions at scale across image and video captioning. On image-captioning benchmarks, an 8B captioner trained with VCap outperforms several leading open- and closed-source MLLMs, including substantially larger systems, achieving strong results on CapMAS and DecapBench. The same training recipe transfers to video captioning, reaching state-of-the-art performance on VDC and ranking competitively on VCapsBench. Human evaluation on a held-out 500-image set ranks VCap first in per-image factual correctness, and the VCap reward agrees with human pairwise preferences on $61.1\%$ of model pairs against a $50\%$ random baseline, providing evidence that the witness-adjudicator signal captures human factual judgments. The resulting captioning checkpoints further generalize to image and video QA without QA-specific fine-tuning, suggesting that VCap improves underlying visual perception rather than only the surface form of captions. Finally, controlled best-of-$N$ and ablation studies show that VCap-driven RL improves beyond what can be explained solely by self-distillation from sampled captions, and that removing either reward dimension or modality produces the failure modes predicted by the two-axis factorization.

We summarize the contributions of this paper as follows:
\begin{itemize}[leftmargin=*]
  \item \textbf{A fact-level framework for visual captioning.} We formalize visual captioning along two complementary axes, Correctness and Completeness, and identify why existing captioning rewards struggle to provide fine-grained supervision for both axes simultaneously.

  \item \textbf{A Witness-Adjudicator reward for weak-to-strong captioning.} We propose VCap, which treats a reference caption as stochastic witness and the visual content as an adjudicator. This role separation yields hypergeometric reward signals, enabling weak-to-strong captioning enhancement.

  \item \textbf{Strong empirical performance and analysis.} Trained with VCap, an 8B captioner achieves leading performance across image- and video-captioning benchmarks. Human evaluation supports its factual correctness, downstream QA results suggest improved visual perception, and controlled best-of-$N$ and ablation studies show that VCap-driven RL goes beyond imitation or self-distillation.
\end{itemize}

\section{Methodology}
\label{sec:method}
This section presents our methodology in three steps. 
First, we formalize visual captioning as a fact-level information extraction task that captures salient visual facts while tolerating lossy representations (Section~\ref{sec:method:prelim}). 
Next, we introduce VCap reward for visual captioning, which assigns complementary roles to reference captions and the visual input, with the former acting as a witness and the latter as an adjudicator (Section~\ref{sec:method:reward}).
Finally, we embed this reward in a combinatorial framework and derive two-axis weak-to-strong generalization directly from hypergeometric expressions (Section~\ref{sec:method:theory}).

\subsection{Formal Definition: Captioning as Fact-Level Lossy Extraction}
\label{sec:method:prelim}

We formalize visual captioning as a lossy, fact-level information extraction problem. Since natural language can only express limited content, captioning is inherently lossy.
A visual signal $x$ (an image or a short video) induces a latent set of describable facts $\mathcal{F}(x)$, which we conceptualize as finite, ranging over objects, attributes, scene events, and relations.  
We denote by $\Phi(y)$ the set of facts expressed by a caption $y$. A reference caption $y_\text{ref}$ induces a fact set $R := \Phi(y_\text{ref})$ which can contain mistakes. For a policy caption $y$, we decompose its fact set as $\Phi(y) = C \cup E$, where $C \subseteq \mathcal{F}(x)$ denotes the set of correct facts supported by $x$, and $E$ denotes the set of facts inconsistent with the visual signal. 
At the fact level, caption quality is governed by two complementary axes: \textbf{Correctness}, requiring every fact in $\Phi(y)$ to be supported by $x$, and \textbf{Completeness}, requiring $\Phi(y)$ to cover as many facts in $\mathcal{F}(x)$ as possible. A reward function for captioning must therefore supervise both axes at the fact level, over a fact space $\mathcal{F}(x)$ that is neither directly observable nor exhaustively enumerable in practice. 
This formulation highlights the central challenge: learning to optimize over an unobserved fact space using only partial and indirect supervision.

\begin{figure}[h]
\centering
\includegraphics[width=\linewidth]{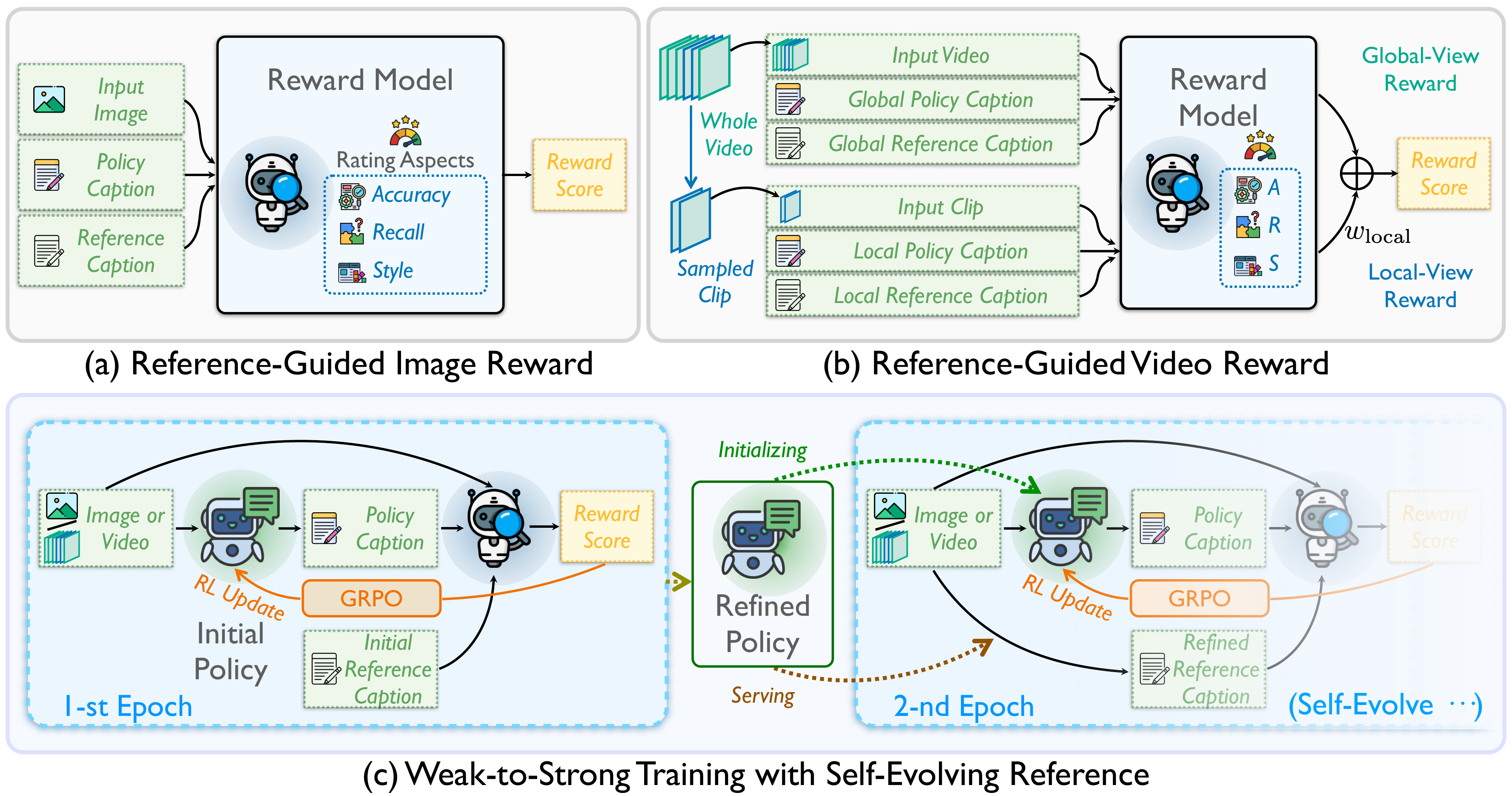}
\caption{\textbf{VCap overview.} (a) VCap's reward mechanism: reference (witness) and image (adjudicator) jointly produce Correctness, Completeness, and Text Quality scores. (b) For video, a global reward and a per‑segment reward are combined. (c) Self‑improvement: the policy model iteratively regenerates stronger references, which sharpen the reward signal to further refine the policy.}
\label{fig:architecture}
\end{figure}

\subsection{VCap: Witness-Adjudicator Reward for Visual Captioning}
\label{sec:method:reward}

As shown in Fig~\ref{fig:architecture} (a), VCap's Witness-Adjudicator Reward is computed by a frozen reward model (MLLM) that takes the triple $(x, y_\text{ref}, y)$ as input and produces a structured judgment in a single evaluation pass.
The reward model is instructed (Appendix~\ref{app:instructions}) to compare $y_\text{ref}$ and $y$ at the fact level, flagging two kinds of problems: facts that are present in $y_\text{ref}$ but missing from $y$ (candidate omissions), and facts in $y$ that contradict $y_\text{ref}$ at a shared slot (candidate hallucinations).
Each candidate problem is then verified against the visual signal $x$: an apparent omission is counted only if $x$ actually supports the missing fact, and an apparent contradiction is counted only if $x$ disagrees with the policy fact (when $x$ disagrees with $y_\text{ref}$ instead, the verdict on $x$ overrides $y_\text{ref}$ so that hallucinations latent in the reference cannot enter the reward).
The reward model lists detected problems and emits three integer scores in $\{0, 1, \dots, 10\}$ summarizing the policy caption: $s_\text{corr}$ measures \textbf{Correctness}, penalizing policy claims that conflict with the reference and are verified by the visual signal. This corresponds to the collision avoidance event $E \cap R = \emptyset$. $s_\text{comp}$ measures \textbf{Completeness}, namely whether the policy caption covers the facts in the reference, with confirmed omissions penalized via the event $R \subseteq C$, and $s_\text{txt}$ measures \textbf{Text Quality} (fluency, coherence, conciseness, and absence of self-evaluative meta-text such as ``all elements have been described,'' so the policy cannot inflate its score by self-assessment).
The three scores are combined into the sentence-level reward $r(x, y_\text{ref}, y)$ by three hyperparameters $(w_\text{corr}, w_\text{comp}, w_\text{txt})$; the explicit aggregation formula is given in Appendix~\ref{app:reward-aggregation}.

For video captioning (Fig~\ref{fig:architecture} (b)), where a long visual signal causes the global pass to miss locally salient facts, we additionally choose a random temporal segment per sample, run the same reward model on the per-segment caption against a per-segment reference to obtain a local reward in exactly the same form, and combine it with the global reward by another fixed weighted sum (also given in Appendix~\ref{app:reward-aggregation}); the local term captures fine-grained temporal facts that a global score blurs, while the global term retains cross-segment events that no single segment can resolve.

In this construction the reference caption and the visual signal play disjoint, non-interchangeable roles.
The reference caption acts as a \textbf{stochastic witness}: its decomposed fact set $R = \Phi(y_\text{ref}) \subset \mathcal{F}(x)$ is read as a uniformly random subset of $\mathcal{F}(x)$ rather than a target to imitate, and serves only to mark the slots of $\mathcal{F}(x)$ at which the policy will be probed.
The visual signal acts as an \textbf{adjudicator}: it never traverses $\mathcal{F}(x)$ as a per-fact oracle, but binarily resolves each fact-level overlap or conflict between $y_\text{ref}$ and $y$ at whichever slot the two meet, with its verdict overriding $y_\text{ref}$ on disagreement.
The two roles are complementary and cannot be swapped: Without visual adjudication, hallucinations latent in $y_\text{ref}$ and superficial wording matches between $y$ and $y_\text{ref}$ would both pollute the reward; without a reference witness, no probability structure exists to localize the policy's facts in $\mathcal{F}(x)$, and the visual content alone can only provide holistic, fact-unlocalized signal. And the supervision they jointly produce is symmetric on both fact-level axes, as the analysis of Section~\ref{sec:method:theory} makes precise.

\subsection{Hypergeometric Analysis: Two-Axis Weak-to-Strong}
\label{sec:method:theory}

By modeling the fact set as a finite combinatorial space, the VCap reward can be analyzed in closed form. We construct that space, express each axis's pass probability as an exact cumulative product, and then deduce all required monotonicity properties without invoking any large‑$N$ approximation.

\paragraph{Finite combinatorial model.}
Let $\mathcal{F}$ denote the latent fact set with $|\mathcal{F}| = N$, viewed as $N$ slots, each holding a distinct fact.
The reference fact set has size $|R| = m$; the policy fact set decomposes as $\Phi(y) = C \cup E$ with $|C| = c$ correct facts and $|E| = n - c$ erroneous facts, so the total caption size is $|\Phi(y)| = n$.
We assume that $R$, $C$, and $E$ are drawn uniformly without replacement from the $N$ slots. All facts are exchangeable and an erroneous fact is detected if and only if it lands on a slot that also belongs to $R$; in that case the visual adjudicator verifies the conflict and confirms the error.
Under these assumptions $R$ supplies a sparse detection net of $m$ armed slots and the visual input acts only on those armed slots.
$N$ is an analytical abstraction, not an algorithmic input: the reward only evaluates the discrete events $R \subseteq C$ and $E \cap R = \emptyset$ via visual adjudication and never estimates $|\mathcal{F}|$.

\paragraph{Completeness axis: subset coverage as a cumulative product.}
The recall test checks whether all $m$ reference facts fall inside $C$, i.e., $R \subseteq C$ (wrong facts cannot cover $R$).
Drawing the $m$ reference facts sequentially without replacement, this probability admits the cumulative‑product expression
\begin{equation}
P_\text{recall}(c \mid m, N)
\;=\; \frac{\binom{N - m}{\,c - m\,}}{\binom{N}{c}}
\;=\; \prod_{i=0}^{m-1} \frac{c - i}{N - i},
\label{eq:p-recall}
\end{equation}
where each factor $(c - i)/(N - i)$ is the conditional probability that the $(i{+}1)$-th reference fact lands in $C$ given that the previous $i$ already have.
When $c < m$, $R \subseteq C$ is impossible and $P_\text{recall} = 0$, providing a strong signal that pushes the policy to increase $c$ until it at least covers the reference facts. Once $c \ge m$, raising $c$ to $c+1$ strictly enlarges every factor, so $P_\text{recall}$ is strictly monotone increasing in $c$ for every $m \ge 1$; the supervision pulls the policy unambiguously toward $c \to N$, the saturation of correct facts at the visual‑information ceiling.
$m$ enters \eqref{eq:p-recall} only as the \emph{number} of factors, not as their location, so enlarging $m$ multiplies more strictly‑increasing‑in‑$c$ factors and hence sharpens the gradient on $c$ but does not move the optimum away from $c \to N$.

\paragraph{Correctness axis: collision avoidance as a cumulative product.}
The collision test asks whether the $n - c$ uniformly drawn error positions all avoid the $m$ reference slots, i.e., $E \cap R = \emptyset$.
Drawing the $n - c$ error positions sequentially without replacement, this probability admits the similar expression
\begin{equation}
P_\text{coll}(n - c \mid m, N)
\;=\; \frac{\binom{N - m}{\,n - c\,}}{\binom{N}{\,n - c\,}}
\;=\; \prod_{j=0}^{n-c-1} \frac{N - m - j}{N - j},
\label{eq:p-coll}
\end{equation}
where each factor $(N - m - j)/(N - j)$ is the conditional probability that the $(j{+}1)$-th erroneous fact misses $R$ given that the previous $j$ already have.
When $n - c = 0$, the empty product equals $1$.
For every $m \ge 1$, each factor is strictly less than one, so $P_\text{coll}$ is strictly monotone decreasing in $n - c$, pulling the policy unambiguously toward $n - c \to 0$ (zero unsupported facts).
A larger $m$ shrinks each ratio and multiplicatively sharpens the penalty for each additional erroneous fact, but does not change the location of the optimum at $n - c \to 0$.

\paragraph{Two-axis weak-to-strong and self-improvement.}
Because $P_\text{recall}$ and $P_\text{coll}$ depend on disjoint variables ($c$ vs.\ $n - c$) and operate on disjoint fact subsets, the reward jointly pushes the policy toward $(c \to N,\; n - c \to 0)$ for every $m \ge 1$, without imposing a structural trade‑off between completeness and correctness.
The reference size $m$ controls only the steepness of supervision (through the number of factors in \eqref{eq:p-recall} and the per‑factor strength in \eqref{eq:p-coll}), never the location of the optimum, so weakening the reference softens the gradient on both axes simultaneously without shifting where the policy is being pushed.
This analysis suggests a natural self-improvement loop. Starting from weak references produced, optimization under the witness-adjudicator reward can push the policy beyond the initial reference quality because the optimum is not tied to the reference itself. If regenerating the references with the trained policy can yield captions that cover more correct facts and contain fewer errors, then the new reference pool has strictly larger average witness size $m$, which by the same equations sharpens the supervision gradient on both axes and accelerates convergence toward the same fixed optimum $(c \to N,\; n - c \to 0)$ in the next iteration (Fig~\ref{fig:architecture} (c)). 
Two‑axis weak‑to‑strong and self‑improvement are therefore two faces of the same fact: under the Witness‑Adjudicator Reward, reference quality controls the speed of convergence and never its destination.

\section{Experiments}
\label{sec:experiments}

We evaluate VCap on the two modalities the closed-form analysis of
Section~\ref{sec:method:theory} predicts it should generalize across, image
captioning and video captioning, and ask three questions.
First, does an 8B captioner trained with the Witness--Adjudicator Reward match
or surpass open- and closed-source state of the art that exceeds it by one to
two orders of magnitude in scale?
Second, do the gains transfer from images to long videos under the same reward,
including the per-segment local term of Equation~\eqref{eq:reward-video}?
Third, does self-improvement through regenerated references yield the gains predicted by the $m$-independence property in Equations~\eqref{eq:p-recall} and~\eqref{eq:p-coll}?
The main-result protocol below answers all three; ablations that decompose
the result along each axis of the closed form follow in
Section~\ref{sec:experiments:ablation}.

\subsection{Setup}
\label{sec:experiments:setup}

We train all VCap variants from \textsc{Qwen3-VL-8B-Instruct} with GRPO,
using the Witness-Adjudicator Reward of Section~\ref{sec:method:reward} on
the triple $(x, y_\text{ref}, y)$, with the per-segment local term of
Equation~\eqref{eq:reward-video} enabled for video; all weights are held
fixed across image and video training (Appendix~\ref{app:rl-setup}).
\textbf{VCap~(e1)} is trained against an initial reference pool from an
off-the-shelf captioner, and \textbf{VCap~(e2)} is retrained from the same
backbone against references regenerated by VCap~(e1), the self-improvement
schedule of Section~\ref{sec:method:theory}.
We evaluate on four benchmarks:
\textbf{CapMAS}~\citep{lee2024toward} (CLAIR, Coverage, Factuality, with Avg
their mean) and
\textbf{DecapBench}~\citep{ye2025painting} (Precision, Recall, DCScore, with
Avg the mean of Precision and Recall) on images, and
\textbf{VCapsBench} (atomic-QA AR/IR/CR) and
\textbf{VDC}~\citep{chai2024auroracap} ($\mathit{tp\_acc}$ on five aspects,
with Avg their mean) on video.
Baselines include the unmodified backbone \textsc{Qwen3-VL-8B-Instruct},
\textsc{Qwen3-VL-235B-Instruct}, \textsc{Qwen3.5-397B},
\textsc{Gemini~3-Flash}, \textsc{Gemini~3.1~Pro}, \textsc{Seed~2.0~Pro},
\textsc{GPT-5.4} (image only), and the caption-RL peer
\textsc{CapRL}~\citep{xing2025caprl} (image only); every baseline is scored
through the identical pipeline.
For all judge-based scores, we evaluate every model using the same pipeline and interface, and strive to remain as consistent as possible with the original paper.

\begin{table}[t]
\centering
\footnotesize
\caption{Image-captioning results on CapMAS and DecapBench.}
\label{tab:image-main}
\setlength{\tabcolsep}{3pt}
\begin{tabular*}{\linewidth}{@{\extracolsep{\fill}} l cccc cccc}
\toprule
& \multicolumn{4}{c}{\textbf{CapMAS}} & \multicolumn{4}{c}{\textbf{DecapBench}} \\
\cmidrule(lr){2-5}\cmidrule(lr){6-9}
Model & CLAIR & Coverage & Factuality & Avg & Precision & Recall & DCScore & Avg \\
\midrule
CapRL                          & 79.65 & 69.53 & 63.76 & 70.98 & 75.93 & 50.09 & 59.47 & 63.01 \\
Qwen3-VL-8B-Instr           & 86.19 & 71.30 & 76.25 & 77.91 & 82.73 & 55.61 & 65.70 & 69.17 \\
Qwen3-VL-235B-Instr    & 84.44 & 71.23 & 77.12 & 77.60 & 79.76 & 55.99 & 64.99 & 67.88 \\
Qwen3.5-397B                   & 84.97 & 73.18 & 77.02 & 78.39 & 78.15 & 56.21 & 64.53 & 67.18 \\
Seed 2.0 Pro                   & 88.03 & 73.43 & 77.68 & 79.71 & 77.74 & 56.95 & 64.89 & 67.35 \\
Gemini 3.1 Pro                 & 89.19 & 72.00 & 78.37 & 79.85 & 83.76 & 56.63 & 66.46 & 70.20 \\
GPT-5.4                          & 88.18 & 73.92 & 81.99 & 81.36 & {86.27} & 60.08 & \underline{70.10} & 73.18 \\
\midrule
\textbf{VCap (e1)}             & 89.14 & \underline{73.95} & \underline{81.48} & \underline{81.52} & \textbf{87.80} & 58.83 & 69.52 & \underline{73.32} \\
\textbf{VCap (e2)}             & \textbf{89.99} & \textbf{74.18} & \textbf{86.42} & \textbf{83.53} & \underline{86.42} & \textbf{60.92} & \textbf{70.35} & \textbf{73.67} \\
\bottomrule
\end{tabular*}
\end{table}

\begin{table}[t]
\centering
\footnotesize
\caption{Video-captioning results on VCapsBench and VDC.}
\label{tab:video-main}
\setlength{\tabcolsep}{1pt}
\begin{tabular*}{\linewidth}{@{\extracolsep{\fill}} l ccc cccccc}
\toprule
& \multicolumn{3}{c}{\textbf{VCapsBench}} & \multicolumn{6}{c}{\textbf{VDC}} \\
\cmidrule(lr){2-4}\cmidrule(lr){5-10}
Model & AR $\uparrow$ & IR $\downarrow$ & CR $\uparrow$ & Background & Camera & Detailed & Main Object & Short & Avg \\
\midrule
Qwen3-VL-8B-Instr           & 63.28          & \underline{11.77}    & 71.73          & 25.63 & 32.54 & 37.51 & 35.28 & 22.96 & 30.78 \\
Qwen3-VL-235B-Instr         & 64.64          & 12.35             & 73.75          & 39.64 & 32.87 & 38.57 & 39.00 & \textbf{23.56} & 34.73 \\
Qwen3.5-397B                   & 65.68          & 12.05             & 74.68          & 37.76 & 30.53 & 35.87 & 36.88 & 22.96 & 32.80 \\
Gemini 3.1 Pro                 & 65.91          & \textbf{11.68}    & 74.62          & 36.56 & 28.54 & 33.98 & 35.43 & 22.57 & 31.41 \\
Seed 2.0 Pro                   & \textbf{76.53} & 13.39             & \textbf{88.36} & 34.91 & 27.39 & 33.54 & 34.55 & 21.54 & 30.39 \\
\midrule
\textbf{VCap (e1)}             & 71.34          & 14.01             & 82.96          & \underline{40.71} & \underline{33.22} & \underline{38.90} & \underline{39.82} & 23.42 & \underline{35.21} \\
\textbf{VCap (e2)}             & \underline{72.15} & 13.67          & \underline{83.57} & \textbf{41.56} & \textbf{34.00} & \textbf{40.22} & \textbf{40.74} & \underline{23.50} & \textbf{36.01} \\
\bottomrule
\end{tabular*}
\end{table}

\subsection{Main Results}
\label{sec:experiments:main}

Tables~\ref{tab:image-main} and~\ref{tab:video-main} report the full results.
On image captioning, \textbf{VCap~(e2) attains the best score on every CapMAS
metric}: CLAIR $89.99$, Coverage $74.18$, Factuality $86.42$, Avg
$83.53$, and on DecapBench Recall ($60.92$), DCScore ($70.35$), and Avg
($73.67$), surpassing \textsc{GPT-5.4} ($81.36$/$73.18$ Avg) and the
open-source \textsc{Qwen3.5-397B} at $30\times$-$50\times$ its parameter
count without any distillation.
On video captioning, \textbf{VCap~(e2) achieves the best VDC score among the evaluated models.} (Avg
$36.01$, $+1.28$ over the next-best \textsc{Qwen3-VL-235B-Instruct}, $+5.23$
over the backbone) on four of the five aspects, and is second on VCapsBench
on both AR ($72.15$) and CR ($83.57$) behind \textsc{Seed~2.0~Pro}
($76.53$/$88.36$); the residual gap on VCapsBench, whose score is dominated
by holistic temporal QA rather than per-fact decomposition, is consistent
with our reward signal matching the metric most tightly when both operate at
the fact level.
\textbf{The e1$\to$e2 self-improvement is monotone on every metric
across all four benchmarks}: CapMAS Avg $+2.01$, DecapBench Avg $+0.35$,
VCapsBench AR $+0.81$, VDC Avg $+0.80$, which is consistent with the $m$-independence prediction of Section~\ref{sec:method:theory}: Regenerating references with the refined
policy sharpens the supervision on both
axes, and pushes the policy toward the visual-information ceiling
without any data curriculum or distillation.

\subsection{Human Evaluation and Reward-Model Alignment}
\label{sec:experiments:human}

\begin{figure}[t]
\centering
\begin{minipage}[c]{0.61\linewidth}
\centering
\footnotesize
\setlength{\tabcolsep}{2.4pt}
\renewcommand{\arraystretch}{1.05}
\begin{tabular}{@{} l rrrr ccc r @{}}
\toprule
Model & $|M|$ & $|I|$ & $\hat{M}$ & $\hat{I}$ & $\bar{r}_\mathrm{H}$ & $\tilde{r}_\mathrm{H}$ & $\bar{r}_\mathrm{V}$ & $\bar{w}$ \\
\midrule
\textbf{VCap (e2)} & \underline{7{,}563}  & \textbf{993}   & \textbf{4{,}137} & \underline{494}  & \textbf{2.27} & \textbf{2} & \textbf{1.46} & 1{,}511 \\
Seed 2.0 Pro       & \textbf{7{,}318} & 1{,}429    & \underline{4{,}289} & \textbf{442}  & 2.85 & 3 & \underline{2.19} & 497 \\
Gemini 3.1 Pro     & 11{,}906          & 1{,}459        & 6{,}964          & 576              & \underline{2.81} & 3 & 3.42 & 446 \\
GPT-5.4              & 13{,}257          & 1{,}851        & 7{,}997          & 996              & 3.58 & 4 & 3.71 & 372 \\
Qwen3.5-397B       & 15{,}684          & \underline{1{,}367} & 9{,}749     & 593              & 3.47 & 4 & 4.21 & 351 \\
\bottomrule
\end{tabular}
\end{minipage}\hfill
\begin{minipage}[c]{0.37\linewidth}
\centering
\includegraphics[width=\linewidth]{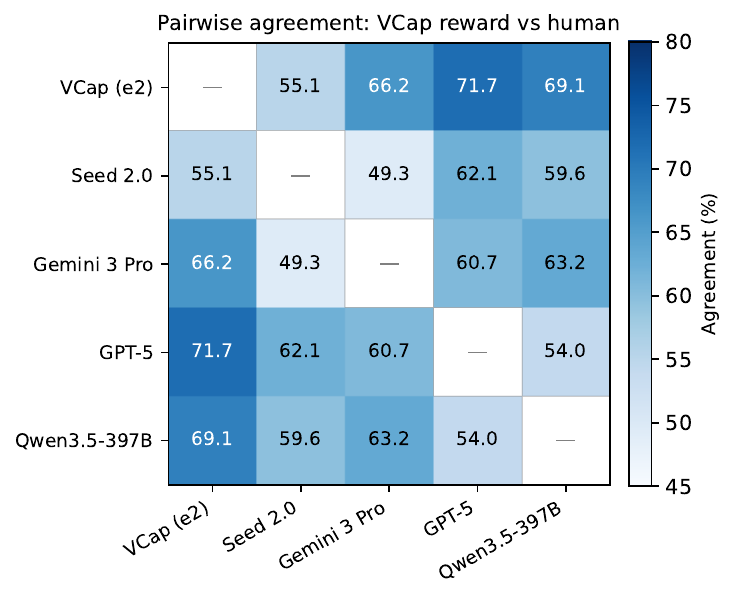}
\end{minipage}
\caption{Human evaluation on the 500-image set.
\textbf{Left:} per-model statistics.
$|M|$/$|I|$: total Judge-proposed missing/inconsistent propositions.
$\hat{M}$/$\hat{I}$: human-confirmed true missing/inconsistent counts.
$\bar{r}_\mathrm{H}$/$\tilde{r}_\mathrm{H}$: mean and median per-image
human rank, where $1$ denotes the best model.
$\bar{r}_\mathrm{V}$: mean per-image rank under the VCap reward.
$\bar{w}$: average caption length in words.
\textbf{Right:} per-image pairwise agreement (\%) between the VCap reward
and human ranking, ties counted as agreement.}
\label{fig:human-eval}
\end{figure}

The benchmark numbers above are produced by automatic judges; we now corroborate
them with human annotation, on the same 500 image set, and use the
human verdicts to test whether the Witness--Adjudicator reward itself ranks
captions the way humans do.
For every image, the five candidate captions, \textsc{VCap~(e2)},
\textsc{Seed~2.0~Pro}, \textsc{Gemini~3.1~Pro}, \textsc{GPT-5.4}, and
\textsc{Qwen3.5-397B}, are mutually used as references, and a Judge model
(\textsc{Qwen3-VL-235B-Instruct}) extracts, for each ordered pair, the
fact-level propositions \emph{missing} from the candidate and those
\emph{inconsistent} with the image; an automatic verifier filters obvious
errors and the surviving propositions are sent to human annotators, who
label each as a true missing/inconsistency or not.
Figure~\ref{fig:human-eval} (left) summarizes the resulting proposition pools
and human-confirmed counts; Figure~\ref{fig:human-eval} (right) reports the
pairwise agreement between the VCap reward and human ranking.

\textbf{Human evaluation confirms VCap as the best image captioner.}
\textsc{VCap~(e2)} ranks first on true missing ($\hat{M}$) and on the human
mean and median per-image rank ($\bar{r}_\mathrm{H}$, $\tilde{r}_\mathrm{H}$),
and is second only to \textsc{Seed~2.0~Pro} on true inconsistencies, despite
producing captions roughly $3$--$4\times$ longer than the closed-source
baselines and therefore exposing far more facts to potential challenge.
\textsc{Seed~2.0~Pro} is the only baseline within $2.2\%$ on total true
issues $\hat{M}{+}\hat{I}$; \textsc{Qwen3.5-397B} produces $2.4\times$ as
many real omissions per caption as \textsc{VCap~(e2)} despite being
$\sim\!50\times$ larger.

\textbf{The Witness-Adjudicator reward agrees with human pairwise
preference.}
For every image we rank the five captions both by the VCap reward and by
the human verdict, and compute, for each unordered pair of models, the
fraction of images on which the two rankings agree on which caption is
better (Figure~\ref{fig:human-eval}, right).
Across all $\binom{5}{2}{=}10$ pairs the mean agreement is
$\mathbf{61.1\%}$ against a random baseline of $50\%$, exceeds $50\%$ on
$9/10$ pairs, and reaches $71.7\%$ on \textsc{VCap~(e2)} vs.\
\textsc{GPT-5.4}; the only pair that dips below random,
\textsc{Seed~2.0~Pro} vs.\ \textsc{Gemini~3.1~Pro} ($49.3\%$), is precisely
the pair humans themselves rank within $0.04$ on $\bar{r}_\mathrm{H}$.
At the macro level, the VCap mean-rank column $\bar{r}_\mathrm{V}$ tracks
$\bar{r}_\mathrm{H}$ closely: both place \textsc{VCap~(e2)} clearly first
and the same two models in the bottom half, disagreeing only on adjacent
near-ties (\textsc{Seed} vs.\ \textsc{Gemini}, \textsc{GPT-5.4} vs.\
\textsc{Qwen3.5}), demonstrating that the reward signal driving training is
well aligned with human judgment.

\subsection{Best-of-$N$ Distillation: VCap Surpasses the Self-Distillation Ceiling}
\label{sec:experiments:bon}

A recent line of work argues RLVR is implicit self-distillation: the
policy converges to its own Best-of-$N$ ceiling and never crosses it.
We test this with three SFT runs that all start from
\textsc{Qwen3-VL-8B-Instruct} and differ only in the teacher whose
BoN-of-$8$ samples (scored by the VCap reward) are used as targets:
the backbone itself (\emph{Self-distill}), \textsc{VCap~(e1)}, and
\textsc{VCap~(e2)}.

\begin{table}[t]
\centering
\footnotesize
\caption{Best-of-$8$ distillation ablation. All students are SFT from
\textsc{Qwen3-VL-8B-Instruct} on the BoN-of-$8$ samples from the listed
teacher, scored by the VCap reward. \textbf{Bold}/\underline{underline}
mark the best/second-best per column.}
\label{tab:bon-distill}
\setlength{\tabcolsep}{3pt}
\begin{tabular*}{\linewidth}{@{\extracolsep{\fill}} l cccc cccc}
\toprule
& \multicolumn{4}{c}{\textbf{CapMAS}} & \multicolumn{4}{c}{\textbf{DecapBench}} \\
\cmidrule(lr){2-5}\cmidrule(lr){6-9}
Model & CLAIR & Coverage & Factuality & Avg & Precision & Recall & DCScore & Avg \\
\midrule
Qwen3-VL-8B-Instr        & 86.19 & 71.30 & 76.25 & 77.91 & 82.73 & 55.61 & 65.70 & 69.17 \\
\quad + Self-distill & 84.84 & 71.09 & 77.02 & 77.65 & 80.98 & 54.95 & 64.86 & 67.97 \\
\midrule
\textbf{VCap (e1)}                     & \underline{89.14}          & \underline{73.95} & 81.48          & 81.52          & \textbf{87.80} & {58.83} & \underline{69.52} & \underline{73.32} \\
\quad + Distill from VCap (e1)     & 88.62          & 73.00             & {84.37} & {82.00} & \underline{86.50}          & 58.56             & 68.71             & 72.53 \\
\midrule
\textbf{VCap (e2)}                     & \textbf{89.99} & \textbf{74.18}    & \textbf{86.42} & \textbf{83.53} & {86.42} & \textbf{60.92}    & \textbf{70.35}    & \textbf{73.67} \\
\quad + Distill from VCap (e2)     & {89.10} & 73.71          & \underline{86.29}          & \underline{83.02}          & 86.22          & \underline{59.75}             & 69.44             & 72.99 \\
\bottomrule
\end{tabular*}
\end{table}

Three observations follow.
(i)~\textbf{Self-distill regresses} ($-0.26$ CapMAS Avg, $-1.20$
DecapBench Avg): the backbone's BoN samples do not introduce new information beyond its existing output distribution; thus naive self-distillation merely re-fits the same manifold with additional sampling noise, challenging the view that the gains from RLVR can be attributed to self-distillation.
(ii)~\textbf{VCap~$e1{,}e2$ lift the same backbone by $+3.6$/$+5.6$
CapMAS and $+4.2$/$+4.5$ DecapBench}, an order of magnitude beyond what
BoN selection extracts.
(iii)~Once VCap has shifted the policy, sample-level distillation
\emph{does} transfer the gain, which means students trained on $e1$/$e2$'s BoN
samples perform close to their teacher on every metric,
and even exceed the $e1$ teacher on CapMAS Avg ($82.00$ vs.\ $81.52$).
The reward, not BoN selection, supplies the shift; sample
imitation suffices only \emph{after} the shift.

\subsection{Generalization Beyond Captioning: Image and Video QA}
\label{sec:experiments:generalization}

If VCap genuinely improves how the policy reads visual facts rather than
merely shaping the caption form, the gain should transfer to downstream
multimodal understanding without any QA supervision.
We test this on (a)~four image-QA benchmarks under the Prism
framework~\citep{qiao2024prism}: AI2D, MMStar, RealWorldQA, V$\star$, and
(b)~four video-QA benchmarks evaluated directly on the trained
checkpoints: MMVU, MLVU, VideoMMMU, and LVBench.
Both VCap variants and all baselines are scored under the identical
pipeline; the same captioning checkpoints used in
Section~\ref{sec:experiments:main} are evaluated here, with no QA-specific
fine-tuning.

\begin{table}[t]
    \centering
    \scriptsize
    \caption{Generalization to image and video QA.
    Image QA applies the Prism framework~\citep{qiao2024prism} on four datasets.
    Video QA reports each benchmark's official metric.}
    \label{tab:vcap-generalization}
    \setlength{\tabcolsep}{3pt}
    \begin{tabular*}{\linewidth}{@{\extracolsep{\fill}} l ccccc cccc c}
    \toprule
    & \multicolumn{5}{c}{\textbf{Image QA (Prism)}} & \multicolumn{5}{c}{\textbf{Video QA}} \\
    \cmidrule(lr){2-6}\cmidrule(lr){7-11}
    Model & AI2D & MMStar & RWQA & V$\star$ & Avg & MMVU & MLVU & VMMMU & LVBench & Avg \\
    \midrule
    CapRL                & 83.71 & 66.87 & 63.27 & 52.36 & 66.56 & --    & --    & --    & --    & --    \\
    Qwen3-VL-8B-Instr    & 81.83 & 65.20 & 60.78 & 50.79 & 64.65 & 60.80 & 74.20 & 57.00 & 45.64 & 60.99 \\
    \midrule
    VCap (e1)            & 83.19 & 66.53 & 60.92 & 48.17 & 64.70 & 62.50 & 74.43 & 58.44 & 46.16 & 63.00 \\
    VCap (e2)            & 83.81 & 67.13 & 62.48 & 52.88 & 66.58 & 63.10 & 73.92 & 60.44 & 46.35 & 63.82 \\
    \midrule
    \textcolor{gray}{Qwen3-VL-235B-Instr} & \textcolor{gray}{85.40} & \textcolor{gray}{69.00} & \textcolor{gray}{62.88} & \textcolor{gray}{52.88} & \textcolor{gray}{67.54} & \textcolor{gray}{70.30} & \textcolor{gray}{80.59} & \textcolor{gray}{72.67} & \textcolor{gray}{54.55} & \textcolor{gray}{69.58} \\
    \textcolor{gray}{Qwen3-VL-235B-Think} & \textcolor{gray}{86.04} & \textcolor{gray}{70.27} & \textcolor{gray}{66.01} & \textcolor{gray}{47.64} & \textcolor{gray}{67.49} & \textcolor{gray}{74.30} & \textcolor{gray}{78.98} & \textcolor{gray}{75.78} & \textcolor{gray}{50.87} & \textcolor{gray}{70.18} \\
    \bottomrule
    \end{tabular*}
\end{table}

\textbf{Caption-only RL transfers to QA.}
On image QA, \textsc{VCap~(e2)} lifts the backbone by $+1.93$ on Prism Avg
($66.58$ vs.\ $64.65$), with consistent gains on every dataset
(AI2D $+1.98$, MMStar $+1.93$, RealWorldQA $+1.70$, V$\star$ $+2.09$), and
\emph{matches} \textsc{CapRL}'s Avg ($66.58$ vs.\ $66.56$) despite
\textsc{CapRL} being trained with a QA-based reward, i.e., on a signal
much closer to the evaluation target than ours.
On video QA, where \textsc{CapRL} is not applicable, the gains are larger:
\textsc{VCap~(e2)} adds $+2.83$ on Avg and $+8.56$ on VideoMMMU over the
backbone, and improves over \textsc{VCap~(e1)} on the more perception-heavy
benchmarks (VideoMMMU $+3.00$, LVBench $+0.20$).
Since VCap was never trained on any QA data, these results suggest that VCap improves visual fact extraction in a way that transfers beyond caption generation, the VCap reward sharpens the policy's
fact-level reading of the visual signal, and that sharper reading is what
both captioning and QA share, validating the broader claim that VCap
strengthens MLLM understanding, not just caption form.

\subsection{Ablation Study}
\label{sec:experiments:ablation}

We isolate the contribution of each input modality and each score
dimension to the Witness--Adjudicator reward by retraining \textsc{VCap (e1)}
under five lesions: removing the reference caption (the reward model
sees only the image), removing the reference image (it sees only the
caption pair), or zeroing one of the three score weights
$w_\text{corr}/w_\text{comp}/w_\text{txt}$ before aggregation.
Table~\ref{tab:ablation} reports image-captioning quality under the same
evaluation pipeline as Section~\ref{sec:experiments:main}.

\begin{table}[t]
\centering
\footnotesize
\caption{Reward-system ablation. ``$-$\,modality'' removes one reference
input from the reward model; ``$-$\,dimension'' zeros the corresponding
score weight before aggregation. All numbers are percentages.}
\label{tab:ablation}
\setlength{\tabcolsep}{3pt}
\begin{tabular*}{\linewidth}{@{\extracolsep{\fill}} l cccc cccc}
\toprule
& \multicolumn{4}{c}{\textbf{CapMAS}} & \multicolumn{4}{c}{\textbf{DecapBench}} \\
\cmidrule(lr){2-5}\cmidrule(lr){6-9}
Setup & CLAIR & Coverage & Factuality & Avg & Precision & Recall & DCScore & Avg \\
\midrule
Qwen3-VL-8B-Instr (no RL)            & 86.19 & 71.30 & 76.25 & 77.91 & 82.73 & 55.61 & 65.71 & 69.17 \\
\midrule
\textbf{VCap (e1) -- full reward}     & 89.14 & 73.95 & 81.48 & 81.52 & 87.80 & 58.83 & 69.52 & 73.32 \\
\midrule
\multicolumn{9}{l}{\emph{Modality ablation}} \\
\quad $-$ reference caption           & 86.45 & 70.24 & 76.27 & 77.65 & 85.80 & 53.24 & 64.82 & 69.52 \\
\quad $-$ reference image             & 87.97 & 73.77 & 81.37 & 81.04 & 86.87 & 56.45 & 67.48 & 71.66 \\
\midrule
\multicolumn{9}{l}{\emph{Dimension ablation}} \\
\quad $-$ Correctness                 & 87.16 & 73.34 & 76.03 & 78.84 & 86.96 & 58.64 & 69.09 & 72.80 \\
\quad $-$ Completeness                & 88.56 & 72.94 & 80.98 & 80.83 & 87.91 & 55.26 & 66.80 & 71.59 \\
\quad $-$ Text Quality                & 88.95 & 74.18 & 81.00 & 81.38 & 87.43 & 58.60 & 69.17 & 73.02 \\
\bottomrule
\end{tabular*}
\end{table}

\textbf{Mixed-modality supervision strictly beats single-modality
supervision.}
Both modality ablations underperform the full reward
($-3.87$/$-0.48$ on CapMAS Avg and $-3.80$/$-1.66$ on DecapBench Avg for
removing reference caption/image, respectively), validating the
non-interchangeable roles assigned to the two reference channels in
Section~\ref{sec:method:reward}.
The asymmetry $-3.87 \!\gg\! -0.48$ on CapMAS Avg, and the analogous gap
on DecapBench, identifies that removing the image effectively disables the adjudicator, yet the damage is milder than removing the witness; this matches our design where the witness first localizes the fact slots to be judged.

\textbf{Per-dimension impact tracks each metric's definition.}
Removing \emph{Completeness} hurts coverage-style metrics most
(Coverage $-1.01$, Recall $-3.57$), while removing \emph{Correctness}
hurts factuality- and precision-style metrics most
(CLAIR $-1.98$, Factuality $-5.45$, Precision $-0.84$), the alignment is
exactly what the construction promises, since $s_\text{comp}$ penalizes
 omissions and $s_\text{corr}$ penalizes contradictions.
Removing \emph{Text Quality}, by contrast, costs only $-0.14$ on CapMAS
Avg and $-0.30$ on DecapBench Avg, the smallest of the three, justifying
its small weight $w_\text{txt}{=}0.01$ in the aggregation.

\section{Conclusion}
\label{sec:conclusion}

We recast visual captioning as fact-level lossy extraction along two independent axes, Correctness and Completeness, and proposed VCap, a Witness Adjudicator reward in which the reference caption serves as a stochastic witness and the image serves as an adjudicator; under this assignment of roles, fact detection follows a hypergeometric distribution, and it is precisely this property that endows VCap with weak-to-strong property, supporting a natural self-improvement loop toward the information ceiling. An 8B captioner trained with VCap surpasses advanced open and closed source SOTAs on visual captioning benchmarks. Alignment with human annotation confirms that VCap supervises both correctness and completeness of captions, the BoN study shows that VCap drives the captioner along a weak-to-strong trajectory that goes beyond what prior work deemed achievable. Generalization experiments verify that VCap broadly enhances the model's visual perception, and ablations further establish the rationality and necessity of VCap's design.

\paragraph{Limitations.}
Four limitations scope our claims. (i)~Image and video policies are trained separately because long-video encoding is roughly an order of magnitude more expensive than image encoding under our hardware budget, so cross-modal synergy under VCap is left open. (ii)~The human evaluation runs on a 500-image set with internal annotators; a larger pool would tighten per-pair agreement and probe long-tail factual phenomena. (iii)~We only optimize with GRPO; whether other on-policy RL algorithms reproduce the same two-axis trajectory is not empirically verified. (iv)~Dedicated captioning specialists comparable to our setting are scarce, undersized, or outdated: among image caption-RL works we only include the recent \textsc{CapRL} as a peer, while video captioning has even fewer up-to-date specialists, forcing the rest of our baselines to lean on closed-source generalist MLLMs.

% {
% \small

% [1] Alexander, J.A.\ \& Mozer, M.C.\ (1995) Template-based algorithms for
% connectionist rule extraction. In G.\ Tesauro, D.S.\ Touretzky and T.K.\ Leen
% (eds.), {\it Advances in Neural Information Processing Systems 7},
% pp.\ 609--616. Cambridge, MA: MIT Press.

% [2] Bower, J.M.\ \& Beeman, D.\ (1995) {\it The Book of GENESIS: Exploring
%   Realistic Neural Models with the GEneral NEural SImulation System.}  New York:
% TELOS/Springer--Verlag.

% [3] Hasselmo, M.E., Schnell, E.\ \& Barkai, E.\ (1995) Dynamics of learning and
% recall at excitatory recurrent synapses and cholinergic modulation in rat
% hippocampal region CA3. {\it Journal of Neuroscience} {\bf 15}(7):5249-5262.
% }
         
  {                                                    
  \small                                                                                                                                                                                                                                   
  \bibliographystyle{plainnat}                        
  \bibliography{reference}                             
  }                                                                           
%%%%%%%%%%%%%%%%%%%%%%%%%%%%%%%%%%%%%%%%%%%%%%%%%%%%%%%%%%%%
\clearpage
\appendix
\section{Appendix}
\label{sec:appendix}

\subsection*{Appendix Contents}
\etocsetnexttocdepth{subsubsection}
\etocsettocstyle{}{}
\localtableofcontents
\bigskip

\clearpage
\subsection{Experimental Setup Details}
\label{app:rl-setup}

This section gathers the implementation details deferred from
Section~\ref{sec:experiments}: the training data behind both image and
video runs (\S\ref{app:training-data}), the hyperparameter settings of the
Witness--Adjudicator Reward (\S\ref{app:reward-aggregation}), and the
on-policy RL algorithm used to optimize it (\S\ref{app:grpo}).

\subsubsection{Training Data}
\label{app:training-data}

\paragraph{Image captioning.}
We use $\sim$29k images sampled from the COCONut dataset, a re-annotated
extension of COCO with high-quality dense panoptic segmentation. COCONut's
broad object inventory and consistent scene coverage make it well suited
for fact-level supervision: any single image typically contains tens of
distinct entities, attributes, and relations, which is the regime in
which the witness--adjudicator construction of
Section~\ref{sec:method:reward} produces the densest signal. The initial
reference pool $\mathcal{D}_\text{ref}(x)$ for VCap~(e1) is generated by the untrained policy backbone \textsc{Qwen3-VL-8B-Instruct} on the same image set; for VCap~(e2) the
pool is regenerated by VCap~(e1) under identical decoding settings.

\paragraph{Video captioning.}
We use $\sim$13k long-form videos collected from YouTube under a topic
distribution that mirrors typical long-video benchmarks (vlogs,
documentary, sports, instructional, tutorials, and scripted scenes).
Videos exceeding a fixed maximum duration are uniformly subsampled at the
frame level to fit the visual context budget of the backbone, but no
segmentation, scripting, or topic filtering is applied. As in the image
case, the initial reference pool is produced by untrained 
\textsc{Qwen3-VL-8B-Instruct}; for the per-segment local term of
Equation~\eqref{eq:reward-video}, we additionally generate a per-segment
reference caption on a random temporal window of each video at every
training step. Reference regeneration for VCap~(e2) follows the same
schedule as in the image case.

\subsubsection{VCap Reward Hyperparameters}
\label{app:reward-aggregation}

\paragraph{Image-level aggregation.}
The three integer scores
$s_\text{corr}, s_\text{comp}, s_\text{txt} \in \{0, 1, \dots, 10\}$
emitted by the reward model in Section~\ref{sec:method:reward} are
combined into the sentence-level reward by a fixed weighted sum with
three hyperparameters $(w_\text{corr}, w_\text{comp}, w_\text{txt})$:
\begin{equation}
r(x, y_\text{ref}, y) \;=\; w_\text{corr}\, s_\text{corr} \;+\; w_\text{comp}\, s_\text{comp} \;+\; w_\text{txt}\, s_\text{txt}.
\label{eq:reward-agg}
\end{equation}
We use $w_\text{corr} = 0.05$, $w_\text{comp} = 0.04$, and
$w_\text{txt} = 0.01$ throughout, held fixed across image and video
training. The asymmetry between the two fact-level weights and the
text-quality weight is deliberate: $s_\text{corr}$ and $s_\text{comp}$
are the two axes whose joint optimum is characterized by the closed-form
analysis of Section~\ref{sec:method:theory}, whereas $s_\text{txt}$ is a
soft regularizer on surface form whose ablation in
Section~\ref{sec:experiments:ablation} costs only $-0.14$ on CapMAS Avg,
justifying its smaller weight. Equation~\eqref{eq:reward-agg} is
intentionally simple: any strictly increasing aggregation preserves the
order of the closed-form landscape derived in
Section~\ref{sec:method:theory} on the two fact-level axes, so the choice
of weights is a hyperparameter on the steepness rather than the location
of the optimum.

\paragraph{Video-level aggregation.}
For video captioning, let
$r_\text{global} = w_\text{corr} s_\text{corr}^\text{g} +
w_\text{comp} s_\text{comp}^\text{g} + w_\text{txt} s_\text{txt}^\text{g}$
denote the reward computed by Equation~\eqref{eq:reward-agg} on the full
video against a global reference, and let $r_\text{local}$ denote the
same form evaluated on a randomly sampled per-segment caption against a
per-segment reference. The video-level sentence reward consumed by the
RL algorithm is
\begin{equation}
r_\text{video}(x, y_\text{ref}, y) \;=\; r_\text{global} \;+\; w_\text{local}\, r_\text{local},
\label{eq:reward-video}
\end{equation}
with $w_\text{local} = 0.1$ held fixed across training. The local term
is intentionally weighted an order of magnitude below the global term:
$r_\text{global}$ is computed against the full video and therefore
carries cross-segment temporal facts that no single window can resolve,
while $r_\text{local}$ is a per-step random snapshot of one window whose
role is to recover locally salient facts that the global pass blurs over
a long visual signal. The $0.1$ ratio supplies this fine-grained signal
without letting it dominate the global fact-level objective.

\subsubsection{GRPO Algorithm}
\label{app:grpo}

We optimize a captioner $\pi_\theta(y \mid x)$ against the sentence-level
scalar reward $r(x, y_\text{ref}, y)$ of Equation~\eqref{eq:reward-agg}
(or Equation~\eqref{eq:reward-video} for video), where $y_\text{ref}$ is
a reference caption drawn from a pool $\mathcal{D}_\text{ref}(x)$
associated with $x$.
The reference pool is initially populated by a moderately strong
off-the-shelf captioner and is later refreshed from $\pi_\theta$ itself in
the self-improvement loop described in Section~\ref{sec:method:theory};
in particular, $y_\text{ref}$ is never assumed to be a high-quality
target to imitate.
Each training step samples a triple $(x, y_\text{ref})$, draws a group of
$G$ candidate captions $\{y^{(i)}\}_{i=1}^{G} \sim \pi_{\theta_\text{old}}(\cdot \mid x)$
from a frozen behavior policy, scores each candidate with
$r(x, y_\text{ref}, y^{(i)})$, and updates $\pi_\theta$ on the resulting
group.

We use GRPO~\citep{shao2024deepseekmath} as the on-policy RL algorithm,
which avoids a learned value baseline by computing per-sample advantages
within the on-policy group:
\begin{equation}
\hat{A}^{(i)} \;=\; \frac{r^{(i)} - \mu_r}{\sigma_r},
\quad
r^{(i)} \;=\; r\bigl(x, y_\text{ref}, y^{(i)}\bigr),
\quad
\mu_r = \mathrm{mean}\!\bigl(\{r^{(j)}\}_{j=1}^{G}\bigr),
\quad
\sigma_r = \mathrm{std}\!\bigl(\{r^{(j)}\}_{j=1}^{G}\bigr),
\label{eq:grpo-adv}
\end{equation}
and maximizes the clipped surrogate objective with a KL anchor on a
frozen reference policy $\pi_\text{ref}$:
\begin{equation}
\mathcal{J}_\text{GRPO}(\theta)
\;=\;
\mathbb{E}\!\left[
\frac{1}{G}\!\sum_{i=1}^{G}\frac{1}{|y^{(i)}|}\!\sum_{t=1}^{|y^{(i)}|}
\Bigl(
\min\!\bigl(\rho_t^{(i)} \hat{A}^{(i)},\; \mathrm{clip}(\rho_t^{(i)}, 1{-}\varepsilon, 1{+}\varepsilon)\,\hat{A}^{(i)}\bigr)
- \beta\, D_\text{KL}\!\bigl(\pi_\theta \,\Vert\, \pi_\text{ref}\bigr)
\Bigr)
\right],
\label{eq:grpo}
\end{equation}
where $\rho_t^{(i)} = \pi_\theta(y_t^{(i)} \mid x, y_{<t}^{(i)}) / \pi_{\theta_\text{old}}(y_t^{(i)} \mid x, y_{<t}^{(i)})$
is the per-token importance ratio, $\varepsilon$ the clipping threshold,
and $\beta$ the KL weight.
The choice of GRPO is orthogonal to our contribution: any on-policy RL
algorithm consuming a sentence-level scalar reward composes with the
Witness--Adjudicator Reward, and the analysis of
Section~\ref{sec:method:theory} applies to the reward landscape
independently of the algorithm.

\subsubsection{Training Configuration}
\label{app:training-config}

All experiments are run on a single $8\times$NVIDIA~H800 node using the
ms-swift framework with vLLM as the rollout sampler and DeepSpeed ZeRO-3
for parameter and optimizer-state sharding. Both the vision encoder and
the vision--language aligner are frozen throughout RL training, so only
the language-model parameters of \textsc{Qwen3-VL-8B-Instruct} receive
gradients. Table~\ref{tab:training-config} lists the optimization,
sampling, and rollout hyperparameters used by the image and video runs.

\begin{table}[t]
\centering
\footnotesize
\caption{Training hyperparameters for image-captioning and video-captioning runs.}
\label{tab:training-config}
\setlength{\tabcolsep}{4pt}
\begin{tabular*}{\linewidth}{@{\extracolsep{\fill}} l c c}
\toprule
Hyperparameter & Image run & Video run \\
\midrule
Hardware                              & 8$\times$H800 & 8$\times$H800 \\
Framework                             & ms-swift      & ms-swift \\
Rollout sampler                       & vLLM          & vLLM \\
Distributed strategy                  & DeepSpeed ZeRO-3 & DeepSpeed ZeRO-3 \\
Precision                             & bfloat16      & bfloat16 \\
Attention                             & FlashAttention & FlashAttention \\
\midrule
Optimizer                             & AdamW         & AdamW \\
Learning rate                         & $5\!\times\!10^{-7}$ & $5\!\times\!10^{-7}$ \\
LR scheduler                          & cosine w/ min-LR & cosine w/ min-LR \\
Min-LR ratio                          & 0.1           & 0.1 \\
Warmup ratio                          & 0.01          & 0.003 \\
Epochs                                & 1             & 1 \\
\midrule
Per-device batch size                 & 2             & 1 \\
Gradient accumulation steps           & 16            & 32 \\
Effective batch size                  & 256           & 256 \\
\midrule
Group size $G$ (\texttt{num\_generations}) & 8        & 8 \\
Sampling temperature                  & 1.0           & 1.0 \\
Top-$p$                                & 0.9           & 0.9 \\
Top-$k$                                & 50            & 50 \\
Repetition penalty                    & 1.05          & 1.05 \\
KL coefficient $\beta$                & $1\!\times\!10^{-3}$ & $1\!\times\!10^{-4}$ \\
\midrule
Max prompt length                     & 14{,}000 tokens & 32{,}000 tokens \\
Max completion length                 & 4{,}096 tokens  & 6{,}300 tokens \\
\bottomrule
\end{tabular*}
\end{table}

\paragraph{Visual encoding budget.}
For image captioning we cap the per-image pixel budget at
$\texttt{max\_pixels} = 6{,}422{,}528$, which corresponds to up to
$\sim$8{,}192 visual tokens per image under the Qwen3-VL patch
tokenizer. For video captioning we sample at most $200$ frames per
video, with each frame allowed to occupy at most $256$ visual tokens
($\le$ 51{,}200 visual tokens per video in the worst case); videos
exceeding the frame budget are uniformly subsampled along the temporal
axis. These caps fit each rollout into the per-modality
\texttt{max\_length} listed in Table~\ref{tab:training-config} (14k for
image, 32k for video) while leaving sufficient headroom for the
generated caption.

\subsection{Related Work}
\label{sec:related}

\subsubsection{Visual Captioning}
\label{app:rw-vcap}

Visual captioning has evolved from short single-sentence descriptions on MS-COCO~\citep{chen2015microsoft} and Flickr30K~\citep{young2014image} toward dense, fact-rich descriptions of images and videos. Recent progress has largely relied on scaling caption supervision through large image-text corpora and synthetic recaptioning pipelines. Early web-scale datasets such as LAION~\citep{schuhmann2022laion}, CC12M~\citep{changpinyo2021conceptual}, and YFCC100M~\citep{thomee2016yfcc100m} enabled large-scale vision-language pretraining, but their alt-text annotations are typically short, noisy, and fact-incomplete.

To improve supervision quality, many works refine or replace noisy captions with stronger synthetic descriptions. BLIP and BLIP-2~\citep{li2022blip,li2023blip} bootstrap captioners from cleaner seed data and use them to relabel web corpora. LaCLIP~\citep{fan2023laclip} rewrites captions with LLM assistance to improve semantic alignment, while CapsFusion~\citep{yu2024capsfusion} consolidates raw alt-text and synthetic captions into denser supervision. A parallel direction distills supervision from frontier proprietary models or human annotators. ShareGPT4V~\citep{chen2024sharegpt4v}, ShareGPT4Video~\citep{chen2024sharegpt4video}, and ALLaVA~\citep{chen2024allava} construct large-scale dense caption datasets using GPT-4V-style systems, while DCI~\citep{urbanek2024picture}, ImageInWords~\citep{garg2024imageinwords}, PixMo-Cap~\citep{deitke2024molmo}, DenseFusion-1M~\citep{li2024densefusion}, and PixelProse~\citep{singla2024pixelprose} further improve caption density through human annotation or curated recaptioning. Similar trends extend to video captioning, including Panda-70M~\citep{chen2024panda70m}, MiraData~\citep{ju2024miradata}, Vript~\citep{yang2024vript}, OpenVid-1M~\citep{nan2024openvid}, and Tarsier2-Recap~\citep{yuan2025tarsier2}.

Another line of work improves caption quality through inference-time composition rather than retraining. These approaches combine object detectors, region-level captioners, or external LLM agents to improve coverage and factual grounding~\citep{wu2024dce,lee2024toward}. While such systems often improve factual coverage, they introduce substantial inference complexity and still ultimately depend on the quality of their constituent captioning modules.

Recent evaluation research has increasingly shifted from holistic similarity metrics toward fact-level assessment. FaithScore~\citep{jing2024faithscore} grounds atomic claims against images, CompreCap~\citep{lu2024benchmarking} models captions as structured scene graphs, and CapMAS~\citep{lee2024toward} evaluates factual correctness and coverage jointly. On video captioning, DREAM-1K~\citep{wang2024tarsier}, VDC~\citep{chai2024auroracap}, CAPability~\citep{liu2025capability}, and ARGUS~\citep{rawal2025argus} similarly decompose evaluation into precision- and recall-oriented dimensions. Standard hallucination benchmarks~\citep{li2023evaluating,wang2023amber,guan2024hallusionbench,sun2023aligning} also remain widely used. Despite this transition toward fact-level evaluation, most caption training objectives still optimize imitation or holistic preference signals, leaving the gap between factual evaluation and factual supervision unresolved.

\subsubsection{Reinforcement Learning for Vision-Language Models}
\label{app:rw-rlvlm}

Reinforcement learning has become a dominant paradigm for post-training large language models. RLHF~\citep{ouyang2022training} demonstrated that preference optimization can substantially improve model behavior beyond supervised fine-tuning, while RL with verifiable rewards (RLVR)~\citep{shao2024deepseekmath,guo2025deepseek} extended this paradigm to domains such as mathematics and code generation. GRPO~\citep{shao2024deepseekmath} has since emerged as a widely adopted optimizer for on-policy RL training.

These ideas have increasingly transferred to vision-language models \citep{contextrl}. Existing approaches can roughly be grouped into three categories. The first category uses task-specific verification signals. For example, CapRL~\citep{xing2025caprl} optimizes caption generation through VQA-based rewards, while related RLVR-style approaches target visual reasoning, grounding, and spatio-temporal perception~\citep{vlmr1,videochatr1}. A second category applies preference optimization or imitation-style objectives to reduce hallucination and improve alignment~\citep{yu2024rlhfv,yu2025rlaifv,xiao2024fgaif,wang2024mdpo,xie2024vdpo}. These methods generally inherit the limitations of the underlying reference or preference signal. A third category moves toward fact-level or set-level reward decomposition. ViCrit~\citep{wang2025vicrit}, SC-Captioner~\citep{sc-captioner2025}, AC-RL~\citep{acrl2025}, RubiCap~\citep{rubicap2025}, and OwlCap~\citep{han2025owlcap} explore rewards based on span localization, precision-recall balancing, rubric scoring, or caption-set equivalence. Related video-captioning approaches include VideoCap-R1~\citep{meng2025videocapr1}, video-SALMONN2~\citep{tang2025videosalmonn2}, VidBridge-R1~\citep{vidbridger12025}, and VDC-Agent~\citep{vdcagent2025}. FeedQuill and DCScore~\citep{ye2025painting} further decompose responses into atomic units and optimize factual precision with length-aware preference signals.

Our work shares the goal of fact-level supervision but differs fundamentally in reward formulation. Existing methods typically treat the reference caption as a target for imitation, semantic matching, or preference comparison, causing the reward optimum to remain coupled to reference quality and coverage. In contrast, VCap treats the reference as stochastic evidence over latent visual facts rather than a gold target. This witness-adjudicator role separation yields a probabilistic factorization over completeness and correctness, producing a reward optimum that remains stable even under weak or incomplete references.

\subsection{Comparison of Reference Roles Across Caption-RL Reward Families}
\label{app:imitation-comparison}

The most informative way to position VCap against prior caption-RL rewards is to ask, for each family, what role the \emph{reference} plays inside the reward computation, and where that role places the optimum.
We compare three representative families against VCap below; in every case the position of the optimum is read off the reward's analytic form rather than from any specific implementation choice.

\paragraph{Imitation-style rewards (BLEU, CIDEr, DPO with a reference target).}
Imitation-style rewards measure $y$ against a fixed reference $y_\text{ref}$ via a similarity function (n-gram overlap, consensus weighting, or pairwise preference toward a reference completion).
In fact-space terms, every such reward is maximized at $\Phi(y) = R$, so its optimum is $c = m$ and $n - c = 0$.
The policy's quality ceiling is therefore the reference's own quality: a reference of size $m$ caps the policy at $c = m$ correct facts and at exactly the facts inside $R$, so a weak reference yields a weak policy by construction, and a noisy reference whose $R \not\subseteq \mathcal{F}(x)$ propagates its hallucinations into the optimum itself.
By contrast, VCap's reward is maximized at $c = N$ and $n - c = 0$ for every $m \ge 1$, with $m$ acting only on the steepness of the climb (Section~\ref{sec:method:theory}).
A reference of size $m = 5$ and a reference of size $m = 50$ therefore agree on \emph{where} the policy should go and disagree only on \emph{how fast}; this is the structural reason why the witness--adjudicator construction breaks the imitation ceiling without any explicit data-curriculum mechanism.

\paragraph{VQA-based rewards.}
VQA-based caption rewards (e.g., CapRL-style pipelines) score a caption by the accuracy with which a downstream QA model can recover answers from it, using a question pool generated from the image as the implicit reference.
On the Completeness side, the question generator is biased toward salient and easily-askable facts, systematically omitting long-tail, relational, and fine-grained facts; the implicit reference $R$ is therefore a non-uniform sample of $\mathcal{F}(x)$, and the optimum location, while approximately $c \to N$ on the questioned subset, is bounded \emph{by question coverage} on the unquestioned tail.
On the Correctness side, hallucinations that the question pool does not interrogate carry no penalty, so the policy can drift in directions the reward cannot see.
VCap removes both pathologies because $R$ is read as a uniform random sample of $\mathcal{F}(x)$ rather than as a question pool, and the image is the source of truth at every armed slot.

\paragraph{VLM-as-judge rewards.}
A VLM-as-judge reward replaces the explicit reference with the judge model's internal memory of $x$, asking the judge to score a caption holistically.
On the Completeness side, the judge attends globally over a long caption, leading to attention drift and high-variance, low-resolution recall signal; on the Correctness side, perceptual errors in the judge become a hidden hallucination channel, since whatever the judge cannot perceive cannot be penalized.
The optimum of such a reward is therefore not described in fact-space terms at all but is determined by judge bias, with no analytic guarantee.
VCap retains the judge's ability to adjudicate at the fact level but strictly localizes its role to the slots armed by an external $R$, which both provides the probability structure that the judge alone lacks and prevents the judge's perceptual blind spots from going untested.

\paragraph{Summary of optimum locations.}
Across the three families above, the location of the optimum is either pinned to the reference's own content (imitation), bounded by the question-pool coverage (VQA), or analytically undefined (VLM-as-judge).
VCap is the only family in which the optimum is fixed at the image-information ceiling $(c \to N,\, n - c \to 0)$ and is decoupled from the reference \emph{content}: Only the reference's geometric position in fact space (its slots, not its answers) matters, and that position controls supervision steepness rather than its location.
This is the analytic statement behind the two-axis weak-to-strong and self-improvement properties of Section~\ref{sec:method:theory}.

\subsection{Video-Captioning Reward-System Ablation}
\label{app:video-ablation}

We mirror the image-captioning ablation of
Section~\ref{sec:experiments:ablation} on the video side: starting from
\textsc{VCap (e1)} we retrain under five lesions, one removing the
reference caption (the reward model sees the video only), three zeroing
one of $w_\text{corr}/w_\text{comp}/w_\text{txt}$ before aggregation,
and a video-specific one zeroing $w_\text{local}$ in
Equation~\eqref{eq:reward-video} so that the per-segment local pass no
longer contributes to the sentence-level reward.
Table~\ref{tab:video-ablation} reports VCapsBench and VDC scores under
the identical evaluation pipeline as Section~\ref{sec:experiments:main}.

\begin{table}[t]
\centering
\footnotesize
\caption{Video reward-system ablation. ``$-$\,modality'' removes the
reference-caption channel from the reward model; ``$-$\,dimension''
zeros the corresponding score weight before aggregation;
``$-$\,local term'' drops the per-segment reward of
Equation~\eqref{eq:reward-video}. All numbers are percentages.}
\label{tab:video-ablation}
\setlength{\tabcolsep}{1pt}
\begin{tabular*}{\linewidth}{@{\extracolsep{\fill}} l ccc cccccc}
\toprule
& \multicolumn{3}{c}{\textbf{VCapsBench}} & \multicolumn{6}{c}{\textbf{VDC}} \\
\cmidrule(lr){2-4}\cmidrule(lr){5-10}
Setup & AR$\uparrow$ & IR$\downarrow$ & CR$\uparrow$ & Background & Camera & Detailed & Main Object & Short & Avg \\
\midrule
Qwen3-VL-8B-Instr (no RL)            & 63.28 & 11.77 & 71.73 & 38.35 & 31.33 & 37.11 & 38.08 & 23.91 & 33.76 \\
\midrule
\textbf{VCap (e1) -- full reward}     & 71.34 & 14.01 & 82.96 & 39.95 & 32.83 & 39.83 & 40.41 & 24.13 & 35.43 \\
\midrule
\multicolumn{10}{l}{\emph{Modality ablation}} \\
\quad $-$ reference caption           & 65.75 & 12.94 & 76.19 & 37.51 & 32.29 & 37.16 & 39.35 & 23.54 & 33.97 \\
\midrule
\multicolumn{10}{l}{\emph{Dimension ablation}} \\
\quad $-$ Correctness                 & 69.14 & 13.21 & 79.66 & 40.71 & 32.42 & 39.59 & 39.82 & 23.76 & 35.26 \\
\quad $-$ Completeness                & 69.41 & 12.16 & 79.02 & 38.75 & 32.72 & 38.90 & 39.53 & 23.56 & 34.69 \\
\quad $-$ Text Quality                & 70.73 & 12.53 & 80.86 & 40.50 & 33.01 & 39.73 & 39.72 & 23.47 & 35.29 \\
\midrule
\multicolumn{10}{l}{\emph{Local-term ablation}} \\
\quad $-$ local term                  & 70.94 & 12.81 & 81.25 & 39.49 & 32.62 & 39.75 & 39.98 & 24.07 & 35.18 \\
\bottomrule
\end{tabular*}
\end{table}

\textbf{The reference caption remains the dominant supervision channel.}
Removing it costs $-5.59$ on VCapsBench AR and $-6.77$ on CR, and
$-1.46$ on VDC Avg---the largest drops among all five lesions. As in
the image case, this confirms the witness role assigned to the
reference caption in Section~\ref{sec:method:reward}: the reward model
extracting facts from the video alone misses many slots that the
witness--adjudicator construction would otherwise mark.

\textbf{Per-dimension impact tracks each metric's definition.}
Zeroing $w_\text{comp}$ produces the larger fact-level recall hit on
both benchmarks (VCapsBench AR $-1.93$, CR $-3.94$; VDC Avg $-0.74$),
while zeroing $w_\text{corr}$ widens the precision gap most clearly on
the holistic VCapsBench measures (AR $-2.20$, CR $-3.30$) and on
fine-grained VDC aspects (Detailed $-0.24$, Main Object $-0.59$).
Zeroing $w_\text{txt}$ costs only $-0.61$ on VCapsBench AR and
$-0.14$ on VDC Avg, the smallest of the three, justifying the
$w_\text{txt}{=}0.01$ weight on the soft surface-form term as in the
image-captioning case.

\textbf{The per-segment local term consistently helps but is small.}
Removing it costs $-0.40$ on VCapsBench AR, $-1.71$ on CR, and
$-0.25$ on VDC Avg---non-trivial but an order of magnitude smaller
than the modality-level lesion, which matches the deliberate
$w_\text{local}{=}0.1$ weight discussed in
Appendix~\ref{app:reward-aggregation}: the local pass recovers
fine-grained per-segment facts that the global pass blurs over the
full video, but the global pass still carries the cross-segment
temporal facts that no single window can resolve.

The IR column rises with the full reward
(\textbf{VCap~(e1)}~$=14.01$ vs.\ backbone $11.77$ and all lesions
$\le 13.21$), echoing the residual gap on VCapsBench discussed in
Section~\ref{sec:experiments:main}: VCapsBench's atomic-QA
construction counts every additional detail-bearing claim as an extra
opportunity to be wrong, so a captioner pushed toward the
information ceiling can exhibit a larger absolute IR even when
fact-level precision improves, as the per-image human evaluation in
Section~\ref{sec:experiments:human} independently verifies.

\tcbset{
  vcapinstrbox/.style={
    breakable,
    colback=cyan!2!white,
    colframe=cyan!55!blue,
    colbacktitle=cyan!22!white,
    coltitle=black,
    fonttitle=\bfseries,
    boxrule=0.6pt, arc=2pt,
    left=5pt, right=5pt, top=4pt, bottom=4pt,
  }
}

\subsection{Instructions for VCap Reward Modeling}
\label{app:instructions}

This subsection collects the four instructions with which the frozen
reward model of Section~\ref{sec:method:reward} is queried: two for
image captioning (Figures~\ref{fig:instr-img-ref}
and~\ref{fig:instr-img-noref}) and two for video captioning
(Figures~\ref{fig:instr-vid-global} and~\ref{fig:instr-vid-segment}).
Each instruction asks the same reward model to (i)~compare the
Generated Caption against the visual signal using the Reference
Caption only as auxiliary information, (ii)~emit a concise analysis
listing confirmed problems, and (iii)~return three integer scores in
$\{0, 1, \dots, 10\}$ inside a single JSON object. To save space, the
second instruction in each pair only spells out the parts that differ
from the first.

\subsubsection{Image Captioning Instructions}
\label{app:instr-img}

\begin{tcolorbox}[vcapinstrbox, title=Instruction 1: Image Caption Reward (with reference)]
\small
You are an expert in image captioning. I will provide a Reference
Caption and a Generated Caption for the Input Image above. Use the
Input Image as ground truth and the Reference Caption as auxiliary
context to evaluate the Correctness, Completeness, and Text Quality of
the Generated Caption.

\textbf{Principles.}
\begin{itemize}
\item The Image is the source of truth. If the Reference Caption
conflicts with the Image, prioritize the Image; the Reference Caption
is auxiliary only.
\item Pay more attention to salient content (main subjects, key
attributes such as count/color/type, actions, crucial spatial
relations) than to background minutiae.
\end{itemize}

\textbf{Operation steps.}
\begin{enumerate}
\item Analyze the Generated Caption on three dimensions:
  \begin{itemize}
  \item \textbf{Correctness}: does it accurately represent the image,
    free of objects not present, inaccuracies, or contradictions?
    Fewer mistakes is better.
  \item \textbf{Completeness}: does it cover all objects in detail
    with no omissions of details or objects? Fewer omissions is
    better.
  \item \textbf{Text Quality}: is it logically fluent, coherent,
    concise (no repetitions), and aesthetically pleasing? Penalize
    any meta-text unrelated to the image content, e.g.\
    self-evaluations such as ``all elements have been described'' or
    ``no detail or object is omitted.''
  \end{itemize}
\item Provide an integer score in $\{0, 1, \dots, 10\}$ for each
dimension.
\end{enumerate}

\textbf{Input.}\\
Reference Caption: \texttt{ref\_answer}\\
Generated Caption: \texttt{gen\_solution}

\textbf{Output format (strict).} Return exactly one JSON object:
\texttt{\{"Analysis": $\langle$your analysis$\rangle$, "Correctness": score1, "Completeness": score2, "Text Quality": score3\}}
\end{tcolorbox}
\captionof{figure}{Reward-model instruction for image captioning when a reference caption is available.}
\label{fig:instr-img-ref}

\paragraph{Reward computation flow.}
At each training step, the placeholders in the appropriate instruction
are filled with the reference caption $y_\text{ref}$
(\texttt{ref\_answer} for images, \texttt{REF\_DESC} for videos) and
the policy caption $y$ (\texttt{gen\_solution} or \texttt{GEN\_DESC});
the no-reference image instruction drops the $y_\text{ref}$ slot
entirely. The filled instruction is submitted to the frozen reward
model together with the visual signal $x$ (the input image, the
timestamp-tagged frames of the full video, or the frames of one
randomly sampled temporal segment, depending on which of the four
instructions is being used) in a single multimodal forward pass. The
reward model returns a JSON-formatted string from which we parse the
three integer scores in $\{0, 1, \dots, 10\}$ along the dimensions
declared by that instruction: $(s_\text{corr}, s_\text{comp},
s_\text{txt})$ for the two image instructions and the per-segment
video instruction, and $(s_\text{reas}, s_\text{corr}, s_\text{comp})$
for the global video instruction, in which case Reasonability occupies
the third slot of the aggregation. The parsed scores are combined into
a sentence-level scalar by the fixed weighted sum of
Equation~\eqref{eq:reward-agg}; for video captioning, the global
reward is further combined with the per-segment local reward by
Equation~\eqref{eq:reward-video} to yield the final scalar consumed by
GRPO.

\begin{tcolorbox}[vcapinstrbox, title={Instruction 2: Image Caption Reward (no reference, ablation)}]               
\small
You are an expert in image captioning. I will provide a Generated
Caption for the Input Image above. Use the Input Image as the sole
source of truth to evaluate the Correctness, Completeness, and Text
Quality of the Generated Caption.

\emph{Dimension definitions and operation steps are identical to
Instruction~1, with two differences:} (i)~no Reference Caption is
provided, so the cross-check between the two captions is dropped and
the analysis grounds directly in the image; (ii)~the Text Quality
criterion additionally penalizes courtesy text, modification notes,
personal feedback for suggestions, and self-evaluations such as ``all
the elements of the image have been clearly described'' or ``no
detail or object is omitted/overlooked''.

\textbf{Input.}\\
Generated Caption: \texttt{gen\_solution}

\textbf{Output format.} Same JSON object as Instruction~1.
\end{tcolorbox}
\captionof{figure}{Reward-model instruction for image captioning without a reference caption, used in the no-reference ablation of Section~\ref{sec:experiments:ablation}.}
\label{fig:instr-img-noref}

\subsubsection{Video Captioning Instructions}
\label{app:instr-vid}

\begin{tcolorbox}[vcapinstrbox, title=Instruction 3: Video Caption Reward (global)]
\small
You are an expert video-description quality evaluator. The frames
above are extracted from a video, each tagged with a timestamp (in
seconds). You are given a Reference Description as auxiliary
information and a Generated Description to be evaluated. Use the
video frames as the \textbf{only} ground truth and rate the Generated
Description on three dimensions: Reasonability, Correctness, and
Completeness.

\textbf{Principles.}
\begin{itemize}
\item Be evidence-based: only contents visible in the frames count as
ground truth. If the Reference Description and the Generated
Description disagree, judge by the frames.
\item Pay main attention to salient content (main subjects, key
attributes, key actions/events over time, important
interactions/spatial relations); more details are preferred.
\end{itemize}

\textbf{Definitions.}
\begin{itemize}
\item \textbf{Reasonability (segmentation)}: are the segment splits
appropriate, with each segment having a coherent and relatively
independent topic, and with boundaries aligned to visible content
changes?
\item \textbf{Correctness}: does the Generated Description avoid
hallucinations, inaccuracies, contradictions, and timestamp
mismatches in each segment's description and in the global summary?
Are the time ranges of important events labelled accurately?
\item \textbf{Completeness}: does it cover all major visible
entities, attributes, fine-grained details, and events, both per
segment and in the global summary?
\end{itemize}

\textbf{Operation steps.}
\begin{enumerate}
\item Read the Generated Description; inspect frames in timestamp
order; compare against the Reference Description and decide by frames
on disagreements.
\item List confirmed hallucinations, inaccuracies, omissions, and
timestamp mismatches per dimension.
\item Provide an integer score in $\{0, 1, \dots, 10\}$ for each
dimension.
\end{enumerate}

\textbf{Input.}\\
Reference Description: \texttt{REF\_DESC}\\
Generated Description: \texttt{GEN\_DESC}

\textbf{Output format (strict).} Return exactly one JSON object:
\texttt{\{"Analysis": $\langle$your analysis$\rangle$, "Reasonability": score1, "Correctness": score2, "Completeness": score3\}}
\end{tcolorbox}
\captionof{figure}{Reward-model instruction for the global pass over a full video, scoring Reasonability, Correctness, and Completeness.}
\label{fig:instr-vid-global}

\begin{tcolorbox}[vcapinstrbox, title=Instruction 4: Video Caption Reward (per-segment local)]
\small
You are an expert video-description quality evaluator. The frames
above belong to a single video \emph{segment}, each tagged with a
timestamp. You are given a Reference Description for the
\emph{whole} video as auxiliary information and a Generated
Description for the segment to be evaluated. Use the segment frames
as the only ground truth and rate Correctness, Completeness, and Text
Quality.

\emph{Principles, evidence-based instruction, and operation steps
follow Instruction~3, with two differences:} (i)~the Reference
Description spans the whole video and contains content outside the
segment; the model is instructed to ignore such out-of-range parts;
(ii)~the Reasonability dimension is dropped (a single segment carries
no segmentation to evaluate) and replaced with \textbf{Text Quality},
which checks fluency, coherence, conciseness (no repetitions),
aesthetic quality, and absence of text unrelated to the video
content.

\textbf{Input.}\\
Reference Description (whole video): \texttt{REF\_DESC}\\
Generated Description (segment): \texttt{GEN\_DESC}

\textbf{Output format (strict).} Return exactly one JSON object:
\texttt{\{"Analysis": $\langle$your analysis$\rangle$, "Correctness": score1, "Completeness": score2, "Text Quality": score3\}}
\end{tcolorbox}
\captionof{figure}{Reward-model instruction for the per-segment local pass on a randomly sampled temporal window, scoring Correctness, Completeness, and Text Quality.}
\label{fig:instr-vid-segment}

\subsection{Case Study: Reference, VCap (e1), and VCap (e2) on the Same Image}
\label{app:case-study}

To make the dual-axis improvement of the Witness--Adjudicator Reward
concrete, we trace a single image from the COCONut training set
through three captioners on the same backbone: (i)~the policy starting
caption from the unmodified \textsc{Qwen3-VL-8B-Instruct}, which we
use here as the reference $y_\text{ref}$ in the witness role,
(ii)~the caption produced by the same backbone after one round of
VCap RL against that reference (\textbf{VCap (e1)}), and (iii)~the
caption produced after a second round in which $y_\text{ref}$ is
regenerated from the VCap (e1) policy itself (\textbf{VCap (e2)},
the self-improvement schedule of Section~\ref{sec:method:theory}).
Across these three captions, fact coverage on Correctness-relevant
content (parked truck, red curb, blue door, overcast sky, etc.) is
preserved while Completeness-relevant content (canopy texture, garage
recess, HVAC vents, flower bed in the bottom-right corner, fabric and
posture details of pedestrians, branch silhouettes, lighting mood)
grows monotonically across $y_\text{ref}\!\to\!\text{e1}\!\to\!\text{e2}$
, which is exactly the trajectory predicted by the closed-form analysis of
Section~\ref{sec:method:theory}, in which a moderately strong
reference sets the steepness of the supervision but never the
location of the optimum, and the self-improvement loop sharpens the
gradient further by raising effective $m$.

\begin{figure}[!htb]
\centering
\includegraphics[width=0.5\linewidth]{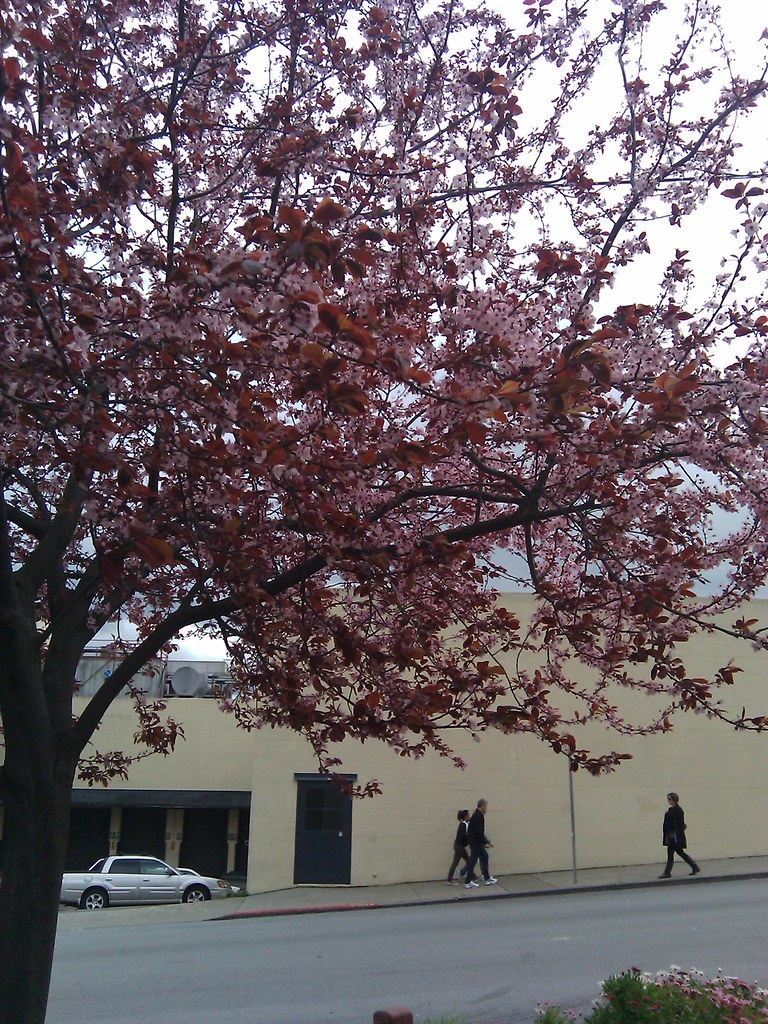}
\caption{Case-study image. A flowering tree with reddish-brown leaves
and pink blossoms dominates the foreground; behind it, three
pedestrians walk along the sidewalk in front of a beige single-story
building with a dark blue door, a recessed garage, and a parked silver
pickup truck. A red-painted curb runs along the road, and a small
flower bed is visible in the bottom-right corner.}
\label{fig:case-study-image}
\end{figure}

\begin{tcolorbox}[vcapinstrbox, title={Reference caption $y_\text{ref}$ -- policy starting caption (\textsc{Qwen3-VL-8B-Instruct}, before RL)}]
\small
This is a street-level photograph capturing a scene of urban
springtime, dominated by a large, flowering tree in the foreground.
The tree, likely a cherry or plum variety, has a thick, dark-barked
trunk and a sprawling canopy of delicate, pinkish-white blossoms
intermingled with reddish-brown leaves, suggesting it might be a
cultivar like the ``Autumnalis'' or ``Kanzan'' cherry. The branches
arch gracefully over the scene, partially obscuring the sky and
casting dappled shadows. The blossoms appear dense and soft, creating
a beautiful contrast against the darker branches and the overcast,
pale sky in the background.

Below the tree, a paved street stretches across the lower part of the
image. A curb painted red runs along the edge of the sidewalk. A
silver pickup truck is parked on the left side of the street, angled
slightly toward the viewer. Behind the truck and the tree, a long,
single-story building with a pale yellow or beige stucco exterior
forms the backdrop. A dark blue or black door with a small window is
visible on the building's facade, and above it, a row of windows or
openings can be seen.

On the sidewalk, three pedestrians are walking from left to right. The
person on the far right is walking alone, wearing a dark coat and dark
pants, and appears to be carrying a bag. To their left, two other
individuals are walking side by side; one is wearing a dark jacket and
jeans, while the other, slightly ahead, is wearing a dark top and
lighter-colored pants. Their movement adds a sense of life and motion
to the scene.

In the bottom right corner, a small patch of greenery with pink and
white flowers can be seen, possibly from a flowerbed or planter,
adding a splash of color to the edge of the frame. The overall
lighting is soft and diffused, typical of an overcast day, which
enhances the pastel tones of the blossoms and the muted colors of the
surroundings. The image evokes a quiet, contemplative moment in a
city neighborhood during the spring season.
\end{tcolorbox}
\captionof{figure}{Policy starting caption from the unmodified backbone, used as the reference $y_\text{ref}$ in this case study.}
\label{fig:case-study-ref}

\begin{tcolorbox}[vcapinstrbox, title={VCap (e1) caption (same backbone after one round of Witness--Adjudicator RL against $y_\text{ref}$)}]
\small
This is an outdoor, street-level photograph capturing a tranquil,
everyday urban or suburban scene framed by the prominent natural
beauty of a flowering tree.

\textbf{Overall Composition and Mood.} The image is framed with a
strong sense of depth and layered elements. The overarching mood is
one of quiet transition---the arrival of spring juxtaposed with the
routine of daily life. The lighting is soft and diffused,
characteristic of an overcast day, which casts even tones across the
scene and enhances the colors and textures without harsh shadows. The
composition is anchored by the massive, sprawling form of the tree in
the foreground, which acts as a natural frame and visual weight,
guiding the viewer's eye toward the background activity.

\textbf{Foreground---the dominant tree.} The tree occupies the entire
left and central portion of the frame, with its trunk visible on the
far left edge and its branches and blossoms cascading across the
upper and middle portions of the image. The visible bark on the left
side appears dark and textured, grounding the composition. The tree
is in full bloom, covered in dense clusters of small, delicate
blossoms in a soft, pastel pink or light lavender hue, creating a
gentle, ethereal wash. Intermingled with the flowers are young,
emerging leaves that are not green but a striking, deep burgundy,
maroon, or reddish-brown---a key visual feature that creates a rich,
warm contrast with the cool pink blossoms and the pale sky. The dark,
intricate network of branches and twigs forms a complex, organic
lacework over the background, partially obscuring and framing the
elements behind it.

\textbf{Foreground---street and sidewalk.} A wide asphalt road
stretches horizontally across the lower third of the image; a narrow,
reddish-pink painted curb line runs parallel to the sidewalk along the
edge of the road. A concrete sidewalk is visible along the base of the
beige building, where the pedestrians are walking.

\textbf{Midground---pedestrians.} Three people walk along the sidewalk
in the midground, moving from right to left. On the far right, a
single person walks alone in dark, likely black, outerwear (a long
coat or jacket) and dark pants; the posture is upright and purposeful.
Closer to the left, two people walk side by side and appear to be
companions: the person on the left wears a dark top and dark pants,
while the person on the right (closer to the building) wears a dark
jacket or top with a noticeable white or light-coloured horizontal
band or trim across the chest/belly area. Both are dressed for cooler
weather, and the spacing of the trio suggests casual, unhurried
movement.

\textbf{Midground---vehicle.} Parked along the left side of the
street, just before the sidewalk begins, is a silver or light gray
pickup truck. Its position is partially obscured by the lower
branches of the large tree and the building's structure, integrating
it into the layered composition.

\textbf{Background---building.} A large, single-story building forms
the dominant background element. Its walls are painted a uniform,
light beige or off-white colour. A distinct, dark blue or black
rectangular door with multiple panels is set into the wall, positioned
centrally in the lower midground. To the left of the door, a dark,
recessed area appears to be a garage entrance or covered walkway;
visible within this recess are vertical structural elements, and
above it, several HVAC units or circular mechanical vents are mounted
on the building's exterior wall. The building's design is functional
and unadorned, serving as a neutral canvas that allows the tree and
the people to stand out.

\textbf{Background---sky.} The sky, visible only through the gaps in
the tree's canopy, is a uniform, pale, overcast white or very light
gray, providing a soft, non-distracting backdrop that allows the
colours of the blossoms and leaves to be prominent.

\textbf{Additional details.} The dominant colours are the muted beige
of the building, the soft pink of the blossoms, the rich burgundy of
the leaves, the dark tones of the trunk, branches and pedestrian
clothing, and the gray of the road and overcast sky---together a
harmonious, subdued palette with accents of warmth from the foliage.
A thin, gray utility pole is visible on the far right side of the
building, just to the right of the central group of pedestrians. In
the bottom right corner, a small portion of a bush or flower bed with
pink and white flowers and green foliage is visible, anchoring the
bottom right and providing a touch of repetition with the tree in the
upper frame.
\end{tcolorbox}
\captionof{figure}{VCap (e1) caption: same backbone after one round of Witness--Adjudicator RL against $y_\text{ref}$.}
\label{fig:case-study-e1}

\begin{tcolorbox}[vcapinstrbox, title={VCap (e2) caption (same backbone after a second round of RL with the reference regenerated from VCap (e1))}]
\small
This image presents a layered urban street scene viewed through the
dominant presence of a flowering tree, creating a composition that
blends natural beauty with everyday human and architectural elements.

\textbf{Overall impression and composition.} The photograph captures
a candid, observational moment on a street corner or sidewalk. The
framing is deliberately angled and partially obscured by the
foreground subject, lending an intimate and slightly voyeuristic
perspective. The overall mood is quiet and contemplative, underscored
by the soft, diffused lighting and the gentle movement of the
pedestrians. There is a harmonious yet contrasting interplay between
the organic, colourful vibrancy of the tree and the muted, structured
geometry of the built environment.

\textbf{Foreground---the dominant tree.} The tree occupies the left
and central portions of the frame, its trunk anchoring the lower-left
corner and its sprawling branches arching diagonally across the upper
half and right side; its sheer size and proximity make it the
undeniable primary subject. The visible trunk is dark, textured, and
substantial, providing a grounding element against the lighter
background, transitioning into a complex network of branches that
extend outward and upward in intricate silhouettes against the sky.
The tree is in a state of unique bloom: leaves and blossoms exhibit a
rich, deep reddish-brown or maroon hue rather than the typical green
or pink, creating a striking, almost autumnal colour statement; the
dense canopy is dotted with soft, pale pink or whitish blossom
clusters nestled among the darker leaves, and the interplay of light
and shadow within the foliage creates depth and texture. The dark,
sinuous branches form strong leading lines that draw the eye inward
and upward, framing the scene behind them and partially veiling the
elements beyond.

\textbf{Midground---pedestrians and infrastructure.} To the right of
the tree's lower branches, three pedestrians walk along the sidewalk,
captured mid-stride; two individuals walk closely together, slightly
ahead of the third, all dressed in dark, casual or business-casual
attire (coats, pants) that contrasts with the warmer tree tones and
blends into the neutral background wall. Their small scale relative
to the tree and the building emphasizes the dominance of the natural
element. Further to the right, a single pedestrian walks alone,
slightly further down the sidewalk---a solitary figure that adds to
the narrative of quiet urban movement.

\textbf{Midground---architecture.} Behind the pedestrians and the
lower branches, a large, plain building wall forms the backdrop,
painted a uniform light beige or pale yellow that provides a neutral
canvas. A dark rectangular door (possibly blue or black) is visible
on the wall, offering a point of contrast in colour and form; to the
left, beneath the overhanging branches, a darker recessed area or
awning structure can be seen at ground level, and higher up near the
roofline behind the upper canopy, faint details that look like HVAC
units or vents are visible, grounding the building in its functional,
real-world context.

\textbf{Midground---vehicle.} Parked along the curb on the left side
of the street, partially sheltered by the tree's lower branches and
the building's recessed area, is a silver or light-coloured pickup
truck whose presence firmly situates the scene in a functional,
accessible urban or suburban setting.

\textbf{Background and atmosphere.} The sky visible through the gaps
in the canopy is a flat, overcast white or very pale gray; this
diffuse lighting eliminates harsh shadows and contributes to the soft,
serene, slightly melancholic atmosphere, enhancing the ethereal
quality of the blossoms and the muted tones of the environment. The
asphalt street occupies the bottom portion of the frame, its gray
surface providing a stable horizontal base, and a distinct red or
reddish-brown painted curb separates the sidewalk from the street,
adding a small but important splash of saturated colour and defining
the boundary of the human realm. At the very bottom right corner, a
hint of additional landscaping is visible---low-lying greenery and a
cluster of small, likely pink or white blossoms---subtly echoing the
main tree's floral theme.

\textbf{Colour palette and visual dynamics.} The scene is defined by
a sophisticated interplay of earthy and muted tones: deep red/brown
in the leaves and blossoms, soft pink/white in the delicate flowers,
beige/yellow in the wall, dark blues/blacks in the door, pedestrian
clothing and tree branches, gray/white in the sky and street, and
red/orange on the curb line. There is strong contrast between the
dark, structural lines of the branches and the lighter background and
between the vibrant foliage and the neutral architecture, but the
overall palette is harmonious due to its earthiness and shared tones.
The image layers multiple textures---rough bark, soft petals and
leaves, smooth wall, asphalt grain, fabric---and combines them with
overlapping elements (branches over building, pedestrians on
sidewalk) to create profound depth and a tangible sense of space.

\textbf{Narrative and mood.} The quiet passage of people against the
backdrop of a uniquely blossoming tree suggests themes of fleeting
moments, seasonal change, and the coexistence of nature within the
constructed urban world. The overcast sky lends a pensive, calm
quality, inviting the viewer to pause and appreciate the subtle
beauty found in an ordinary street corner; the tree acts as both a
shelter and a frame, turning a mundane scene into something quietly
poetic and visually rich, balancing the ephemeral (blossoms, passing
people) with the permanent (building, street).
\end{tcolorbox}
\captionof{figure}{VCap (e2) caption: same backbone after a second round of RL with the reference regenerated from VCap (e1).}
\label{fig:case-study-e2}

\paragraph{What this case study illustrates.}
The reference $y_\text{ref}$ in Figure~\ref{fig:case-study-ref}, which
is also the policy's starting caption, covers the salient facts
(tree, three pedestrians, building, truck, red curb, overcast sky,
flower bed) in roughly $m$ slots, with one mildly speculative
attribution (``Autumnalis or Kanzan cultivar''). After one round of
VCap RL against this same reference, the
\textbf{e1}~caption~(Figure~\ref{fig:case-study-e1}) retains every
Correctness-relevant fact present in $y_\text{ref}$ while
substantially expanding Completeness: the recessed garage, the
HVAC/mechanical vents on the building, the white horizontal trim on
one pedestrian's jacket, the utility pole's position, the layered
canopy/branch lacework, and the explicit colour relations between
blossoms, leaves and sky are all introduced for the first time, and
each is grounded in the image rather than invented. After a second
round in which the reference is regenerated from the e1 policy,
\textbf{e2}~(Figure~\ref{fig:case-study-e2}) preserves the e1 facts
and adds further structural and atmospheric content (the diagonal
arc of the canopy, the leading-line silhouettes of the branches, the
voyeuristic angled framing, the pedestrians' small relative scale,
and an explicit decomposition of the colour palette and texture
layering) without introducing new hallucinations. The trajectory
$y_\text{ref}\!\to\!\text{e1}\!\to\!\text{e2}$ is therefore a monotone
climb on Completeness with no Correctness regression at either step:
a single qualitative instance of the dual-axis
$(c\!\to\!N,\,n-c\!\to\!0)$ optimum predicted in
Section~\ref{sec:method:theory}, and a direct illustration of the
self-improvement loop.

\subsection{Human Annotation Protocol}
\label{app:human-annot}

This subsection details the annotation protocol used to produce the
human verdicts of Section~\ref{sec:experiments:human}. Each annotation
unit is a single image accompanied by five candidate captions
(\textsc{VCap (e2)}, \textsc{Seed 2.0 Pro}, \textsc{Gemini 3.1 Pro},
\textsc{GPT-5.4}, \textsc{Qwen3.5-397B}); for every caption the Judge
model (\textsc{Qwen3-VL-235B-Instruct}) emits two proposition lists:
\emph{missing propositions}---statements asserting that some image
content is absent from the caption---and \emph{inconsistent
propositions}---statements asserting that some content described in the
caption disagrees with the image. Annotators decide each proposition in
two steps and then produce three rankings over the five captions.

\paragraph{Step~1: proposition vs.\ image.}
The annotator compares the proposition only against the image; the
caption is not consulted at this stage. The annotator iterates between
the proposition and the image (zooming, scanning the regions referenced
by the proposition) until reaching one of three verdicts:
\texttt{1.~consistent} (the image actually contains what the
proposition asserts about it), \texttt{2.~inconsistent} (the image
clearly contradicts the assertion), or \texttt{3.~undecidable} (the
referenced object is absent, or visible but not resolvable to the
required level of detail due to occlusion, low resolution, or
ambiguous angle). For an inconsistent proposition with the form ``the
image actually shows $A$, but the caption says $B$,'' Step~1 only
adjudicates the ``image actually shows $A$'' clause.
\texttt{3.~undecidable} is reserved for genuinely unresolvable cases:
whenever the relevant object is visible and clear, the annotator must
commit to either consistent or inconsistent.

\paragraph{Step~2: proposition vs.\ caption (conditional).}
Step~2 is executed only when Step~1 returns \texttt{1.~consistent};
otherwise it is left blank. The annotator first locates the relevant
object in the image, then locates the corresponding span in the caption
(most captions follow a foreground/midground/background or
left/center/right organization), and compares the proposition, the
caption span, and the image jointly. The verdict is one of:
\texttt{1.~not~holding} (the proposition does not in fact hold against
the caption: a missing proposition whose content is already covered by
the caption via a synonym or hypernym, or an inconsistent proposition
whose claimed wording $B$ either is not present in the caption or is
weaker than the caption's actual wording, which is closer to the
image), \texttt{2.~holding} (the proposition genuinely holds: the
caption indeed omits the content, or indeed contains a description that
disagrees with the image), and \texttt{3.~ambiguous} (the relevant
caption span is phrased so vaguely that the verdict cannot be
determined). For missing propositions, annotators must read the entire
caption (not only the locally matched span) before committing to
\texttt{2.~holding}, because models frequently distribute related
content across non-adjacent paragraphs. If a caption self-contradicts,
the existence of the erroneous wording is sufficient to label
\texttt{2.~holding} on the inconsistent side.

\paragraph{Step~3 per-image rankings.}
After all propositions for the five captions of an image are annotated,
the annotator produces three independent rankings over the captions:
\emph{accuracy} (fewer image-disagreeing errors), \emph{completeness}
(fewer omissions), and \emph{granularity} (richer detail and
finer-grained coverage). Each ranking is a strict permutation of
$\{1,\dots,5\}$ with no ties; ties are broken by the annotator's
overall judgment. Accuracy and completeness are anchored to the Step-2
\texttt{2.~holding} counts from inconsistent and missing propositions
respectively, while granularity is read directly off the caption text.
Rankings are produced once per image and not retroactively edited.

\paragraph{Scope and consistency rules.}
Annotation is restricted to objective, image-grounded facts: objects,
attributes, counts, positions, on-image text, colors, actions, and
spatial relations. Subjective content---atmosphere, mood, aesthetic
commentary, narrative interpretation, photographic-intent
analysis---is out of scope; if a proposition lands on subjective text
it is labeled \texttt{3.~undecidable} at Step~1 with the note
``subjective.'' Synonyms and hypernyms count as ``mentioned'' at the
entity level, but attributes (color, count, model) must match at their
own granularity. Each proposition is decided independently of its
neighbors and is not reused across captions. Label codes follow the
exact \texttt{1./2./3.}~prefix format above so that the labels parse
downstream.

\paragraph{Quick reference.}
Table~\ref{tab:human-annot-quick} summarizes the most common
proposition--image--caption configurations and the (Step~1, Step~2)
labels they map to, mirroring the cheat sheet given to annotators.

\begin{table}[t]
\centering
\footnotesize
\caption{Quick reference for the two-step proposition labels used in the
human annotation protocol.}
\label{tab:human-annot-quick}
\setlength{\tabcolsep}{4pt}
\begin{tabular*}{\linewidth}{@{\extracolsep{\fill}} p{0.6\linewidth} c c}
\toprule
Configuration & Step~1 & Step~2 \\
\midrule
Image really contains the asserted content; caption indeed omits it & \texttt{1.~consistent} & \texttt{2.~holding} \\
Image really contains the asserted content; caption already covers it via a synonym/hypernym & \texttt{1.~consistent} & \texttt{1.~not~holding} \\
Image really contains the asserted content; caption is too vague to decide & \texttt{1.~consistent} & \texttt{3.~ambiguous} \\
Image does not contain the asserted content & \texttt{2.~inconsistent} & --- \\
Object is occluded / blurred / unresolvable in the image & \texttt{3.~undecidable} & --- \\
Proposition: ``image shows $A$''; image is indeed $A$; caption says $B$ & \texttt{1.~consistent} & \texttt{2.~holding} \\
Proposition: ``image shows $A$''; image is indeed $A$; caption also says $A$ (proposition misjudged) & \texttt{1.~consistent} & \texttt{1.~not~holding} \\
Proposition: ``image shows $A$''; image actually shows $C$ ($\neq A,\neq B$) & \texttt{2.~inconsistent} & --- \\
\bottomrule
\end{tabular*}
\end{table}

\end{document}